\documentclass[journal]{IEEEtran}
\usepackage{amsmath,amssymb,amsfonts}
\usepackage{algorithmic}
\usepackage{graphicx}
\usepackage{textcomp}
\usepackage{xcolor}
\usepackage{glossaries}
\usepackage{tabularx,ragged2e}
\usepackage{booktabs}
\usepackage{subfloat}
\usepackage{subcaption}
\usepackage{array,multirow,graphicx}
\usepackage{float}
\usepackage{mathtools}
\usepackage{footmisc}
\usepackage{xcolor}
\usepackage{balance}

\newcommand{\rev}[1]{\textcolor{black}{#1}}

\newcommand{\I}{\mathbf{I}}
\newcommand{\Ic}{\mathbf{I}e^{j \mathbf{\Phi}}}
\newcommand{\Ici}{\mathbf{I}_ie^{j \mathbf{\Phi}_i}}
\newcommand{\Iideal}{\mathbf{I}_0 e^{j \mathbf{\Phi}_{0}}}
\newcommand{\F}{\mathbf{F}}
\newcommand{\FH}{\mathbf{H}}
\newcommand{\FHg}{\mathbf{H}_G}
\newcommand{\FHgmax}{\mathbf{H}_{G_\textrm{max}}}
\newcommand{\FHrmax}{\mathbf{H}_{R_\textrm{max}}}
\newcommand{\FHdmax}{\mathbf{H}_{D_\textrm{max}}}
\newcommand{\FHd}{\mathbf{H}_D}
\newcommand{\FHr}{\mathbf{H}_R}

\newcommand{\M}{\mathbf{M}}

\newcommand{\DR}{\mathcal{D}}
\newcommand{\TR}{\mathcal{T}}

\newcommand{\Pbar}{\bar{\mathbf{M}}}
\newcommand{\Ibar}{\bar{\mathbf{I}}}
\newcommand{\Ibard}{\bar{\mathbf{I}}_0}
\newcommand{\Ibarsc}{\bar{\mathbf{I}}_{S}e^{j \Phi_{S}}}
\newcommand{\Ibarh}{\bar{\mathbf{I}}_{H}}

\newcommand{\Ir}{\mathbf{I}_{R}}

\newcommand{\Irc}{\mathbf{I}_{R} e^{j \mathbf{\Phi}_{R}}}
\newcommand{\Itilde}{\tilde{\mathbf{I}}}
\newcommand{\Ptilde}{\tilde{\mathbf{M}}}
\newcommand{\Sa}{\mathbf{S}}
\newcommand{\h}{\mathbf{h}}
\newcommand{\Sn}{\mathbf{S}e^{j \mathbf{\Phi}_{S}}}

\newacronym{dl}{DL}{Deep Learning}
\newacronym{cnn}{CNN}{Convolutional Neural Network}
\newacronym{dncnn}{DnCNN}{Denoising Convolutional Neural Network}
\newacronym{roc}{ROC}{Receiver Operating Characteristic}
\newacronym{auc}{AUC}{Area Under the Curve}
\newacronym{ba}{BA}{Balanced Accuracy}
\newacronym{tpr}{TPR}{True Positive Rate}
\newacronym{fpr}{FPR}{False Positive Rate}
\newacronym{sar}{SAR}{Synthetic Aperture Radar}
\newacronym{ft}{FT}{Fourier Transform}
\newacronym{ift}{IFT}{Inverse Fourier Transform}
\newacronym{st}{ST}{Staring SpotLight}
\newacronym{slc}{SLC}{Single Look Complex}
\newacronym{mlc}{MLC}{Multi Look Complex}
\newacronym{ssc}{SSC}{Single Look Slant Range Complex}
\newacronym{psnr}{PSNR}{Peak Signal to Noise Ratio}
\newacronym{ssim}{SSIM}{Structural Similarity Index Measure}
\newacronym{mssim}{MS-SSIM}{Multi-Scale Structural Similarity Index Measure}
\newacronym{enl}{ENL}{Equivalent Number of Looks}
\newacronym{sae}{SAE}{SAR Adapted Extractor}
\newacronym{asae}{ASAE}{Augmented SAR Adapted Extractor}
\newacronym{asaed}{ASAE-D}{ASAE-Downscaling}
\newacronym{asaeud}{ASAE-UD}{ASAE-UpDownscaling}
\newacronym{mstar}{MSTAR}{Moving and Stationary Target Acquisition and Recognition}
\newacronym{sid}{SID}{Spectral Information Divergence}
\newacronym{sam}{SAM}{Spectral Angle Mapper}
\newacronym{fmse}{FMSE}{Frequency Mean Squared Error}
\newacronym{nfmse}{NFMSE}{Normalized \gls{fmse}}
\newacronym{grd}{GRD}{Ground Range Detected}
\newacronym{prnu}{PRNU}{Photo Response Non Uniformity}
\newacronym{esa}{ESA}{European Space Agency}
\newacronym{fans}{FANS}{Fast Adaptive Non-local SAR Despeckling}
\newacronym{nlsar}{NL-SAR}{Non-Local framework for (In)(Pol)SAR denoising}
\newacronym{truforsar}{TruForS}{TruForSAR}
\newacronym{iou}{IoU}{Intersection Over Union}
\newacronym{f1}{F1}{F1 score}
\newacronym{atr}{ATR}{Automatic Target Recognition}

\begin{document}

\sloppy

\title{Hiding Local Manipulations on SAR Images: \\ a Counter-Forensic Attack}

\author{Sara~Mandelli,~\IEEEmembership{Member,~IEEE,}
        Edoardo~Daniele~Cannas,~\IEEEmembership{Student~Member,~IEEE,}
        Paolo~Bestagini,~\IEEEmembership{Member,~IEEE,}
        Stefano~Tebaldini,~\IEEEmembership{Member,~IEEE,}
        and~Stefano~Tubaro,~\IEEEmembership{Senior~Member,~IEEE}
\thanks{Authors are with the Dipartimento  Elettronica, Informazione e Bioingegneria - Politecnico di Milano - Milan, Italy (email: name.surname@polimi.it).}
\thanks{This material is based on research sponsored by the Defense Advanced
Research Projects Agency (DARPA) and the Air Force Research Laboratory
(AFRL) under agreement number FA8750-20-2-1004. 
The U.S. Government is authorized to reproduce and distribute reprints for Governmental purposes
notwithstanding any copyright notation thereon. The views and conclusions
contained herein are those of the authors and should not be interpreted as necessarily representing the official policies or endorsements, either expressed
or implied, of DARPA and AFRL or the U.S. Government. 
This work was supported by the FOSTERER project, funded by the Italian Ministry of Education, University, and Research within the PRIN 2022 program.
This work was partially supported by the European Union - Next Generation EU under the Italian National Recovery and Resilience Plan (NRRP), Mission 4, Component 2, Investment 1.3, CUP D43C22003080001, partnership on “Telecommunications of the Future” (PE00000001 - program “RESTART”) and by the Investment 1.3, CUP D43C22003050001, partnership on ``SEcurity and RIghts in the CyberSpace’’ (PE00000014 - program ``FF4ALL-SERICS’’).}
}


\maketitle

\begin{abstract}
The vast accessibility of \gls{sar} images through online portals has propelled the research across various fields. This widespread use and easy availability have unfortunately made \gls{sar} data susceptible to malicious alterations, such as local editing applied to the images for inserting or covering the presence of sensitive targets. 
To contrast malicious manipulations, 
in the last years the forensic community has begun to dig into the \gls{sar} manipulation issue, proposing detectors that effectively localize the tampering traces in amplitude images. 
Nonetheless, in this paper we demonstrate that an expert practitioner can exploit the complex nature of \gls{sar} data to obscure any signs of manipulation within a locally altered amplitude image. 
We refer to this approach as a counter-forensic attack.
To achieve the concealment of manipulation traces, the attacker 
can simulate a re-acquisition of the manipulated scene
by the \gls{sar} system that initially generated the pristine image.
In doing so, the attacker can obscure any evidence of manipulation, making it appear as if the image was legitimately produced by the system. 
\rev{This attack has unique features that make it both highly generalizable and relatively easy to apply. First, it is a black-box attack, meaning it is not designed to deceive a specific forensic detector. Furthermore, it does not require a training phase and is not based on adversarial operations.}
We assess the effectiveness of the proposed counter-forensic approach across diverse scenarios, examining various manipulation operations. The obtained results indicate that our devised attack successfully eliminates traces of manipulation, \mbox{deceiving even the most advanced forensic detectors.} 
\end{abstract}

\IEEEpeerreviewmaketitle

\vspace{-15pt}
\section{Introduction}
\label{sec:intro}
\glsreset{sar}

\gls{sar} images are remote sensing data that use radar waves to create two-dimensional or three-dimensional representations of objects or landscapes. 
Over the last years, many online portals have given access to \gls{sar} images in easy-to-download and manageable data products~\cite{copernicushub, iceye, capella}. 
This phenomenon bolstered the research in the \gls{sar} field, with various applications in geology, hydrology, forestry, agriculture, urban planning, disaster management, intelligence, and military investigations~\cite{Moreira2013, Tsokas2022sar}.
For example, \gls{sar} data can localize sensitive targets like army vehicles~\cite{Hummel2000model}, ships~\cite{Chang2019ship}, airports~\cite{Chen2020new} and aircrafts~\cite{Wang2016object}.
\begin{figure}[t]
  \centering  \includegraphics[width=\columnwidth]{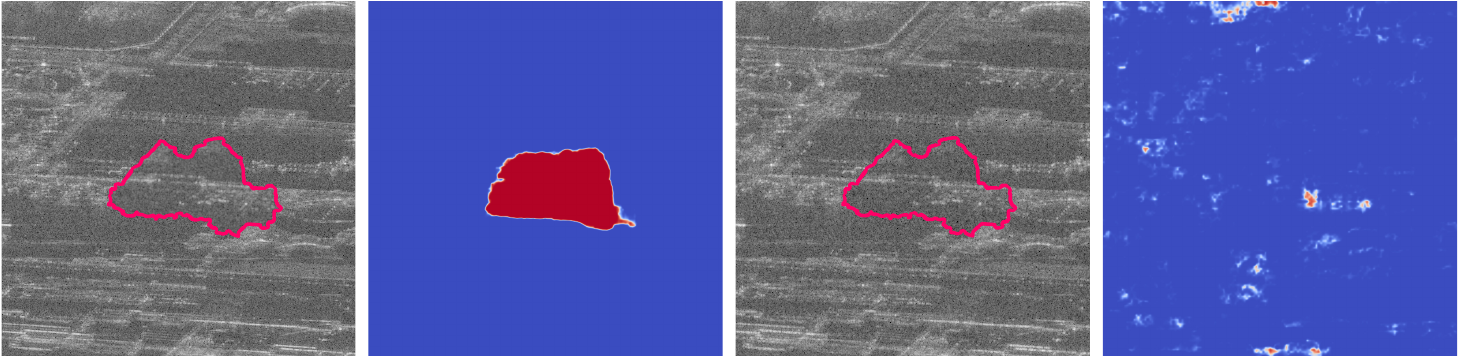}
  \caption{High-level description of the paper goal. Given a manipulated SAR amplitude image (left), state-of-the-art SAR forensic detectors estimate a \rev{real-valued heatmap} (second column) that highlights local image inconsistencies. Our proposed counter-forensic attack simulates a SAR re-acquisition of the manipulated image. The attack leaves untouched the semantic content of the image (third column), but it conceals its tampering traces, such that the estimated \rev{heatmap} (right) does not allow anymore to localize the forgery. The manipulation area is surrounded by the red contour. }
  \label{fig:intro_example}
\vspace{-10pt}
\end{figure}

Unfortunately, the vast popularity of \gls{sar} images and the presence of easy-to-manage products make them a target for malicious manipulations. Despite these images being complex signals, most \gls{sar} data products only provide amplitude information, i.e., they report information as a real-valued data matrix. 
An attacker, even an inexperienced one, can open these images, edit them with a common editing software
and alter their content~\cite{cannas2022amplitude}. 
For instance, inserting or covering the presence of military vehicles is becoming a relatively easy task and has the potential of carrying severe consequences in the context of ground monitoring analysis. 

For this reason, \gls{sar} images are becoming objects of investigation by the multimedia forensic community~\cite{abady2022overview, cannas2022amplitude, cannas2023deep}. 
Given the particular processing pipeline of \gls{sar} images, researchers have developed techniques tailored explicitly to them~\cite{abady2022overview}, proposing solutions for the localization of tampered-with pixel regions~\cite{cannas2022amplitude}.
An example of the \rev{localization heatmap} returned by \rev{one state-of-the-art forensic detector} is shown in Fig.~\ref{fig:intro_example}. Given a manipulated image (left), the detector is able to localize the tampering area with high accuracy (second column). 

So far, the forensic community has worked considering the amplitude information of \gls{sar} signals exclusively. However, \gls{sar} signals are naturally complex, i.e., they report both amplitude and phase. Phase information is crucial in specific applications, such as interferometry~\cite{Moreira2013}, and makes \gls{sar} images sensibly different from standard digital photographs or other modalities of remote sensing data. These differences emerge in the acquisition pipeline of \gls{sar} data and some unique degradation phenomena that afflict it~\cite{bamler1993sar, cannas2022amplitude}. One is speckle noise, i.e., the coherent interference of the scattered waves from multiple targets within a resolution cell~\cite{goodman1976speckle}. 

In this work, we prove that an experienced practitioner can exploit the complex nature of \gls{sar} data to conceal the manipulation traces from a locally manipulated \gls{sar} amplitude image. 
If any information on the \gls{sar} satellite used to acquire the original image is available, it can be 
used to simulate a complete re-acquisition of the manipulated scene. 
In doing so, we show that it is possible to produce an image with the same semantic content of the manipulated one, but with similar \gls{sar} characteristics to pristine ones, where the tampering traces have been hidden.
A high-level example of the tackled goal is shown in Fig.~\ref{fig:intro_example}. Given a manipulated amplitude \gls{sar} image (left), we produce a modified version of it (third column)
such that the \rev{heatmap} estimated by the forensic detector does not highlight the forgery, but resembles that of a pristine image, hiding the manipulation artifacts (right). 

The forensic community refers to this kind of operation as a counter-forensic attack. 
The goal of such attacks is usually twofold. On the one hand, the attack simulates the scenario in which a user \rev{first performs a manipulation and then seeks to conceal the traces of it}. Notice that this does not involve malicious intents only; indeed, it can answer to privacy preserving needs in case specific image manipulations are inserted to hide sensitive targets. In this scenario, the attack allows to avoid the manipulation traces to be exposed. 
On the other hand, the attack allows to test the robustness of existing forensic detectors in spotting the manipulation even in presence of counter-forensic operations. 

\rev{In the literature, the counter-forensic attacks developed to fool tampering localization detectors are all tailored to standard digital pictures~\cite{cao2024transferable, fang2023attacking, boato2024adversarial, rozsa2020adversarial, tailanian2024diffusion}. They often rely on the knowledge of the targeted detector’s functioning~\cite{cao2024transferable, fang2023attacking, boato2024adversarial, rozsa2020adversarial} and can require access to the distribution of pristine samples~\cite{ fang2023attacking, rozsa2020adversarial}. Moreover, when considering deep-learning-based methods, these techniques require training on datasets with characteristics similar to those of test data~\cite{cao2024transferable, fang2023attacking}. Our proposed attack differs sensibly from these methodologies: first, it consists of a black-box attack, i.e., it is not tailored to deceive a specific forensic detector; second, it does not require a dataset of pristine and/or manipulated samples for training or performing adversarial operations.}

In our proposed counter-forensic attack, we alter the amplitude component of a complex \gls{sar} product, and we show that we are able to conceal the forensic footprints left by the manipulation by modeling the \gls{sar} acquisition and processing system with different strategies.
\rev{We purposely avoid any training pipeline, as we aim at simulating a scenario in which an attacker may have little knowledge of the \gls{sar} system used for acquiring the image they want to attack.}

We build a dataset of manipulated \gls{sar} amplitude images and validate the proposed pipeline in various scenarios, considering different manipulation operations and investigating in depth the effect of our methodology on their traces. 
The achieved results show that the proposed attack wipes out the manipulation traces, fooling the most advanced forensic detectors \rev{and outperforming other counter-forensic solutions}. Moreover, the produced images carry typical \gls{sar} characteristics similar to original pristine ones. 

To summarize, our contributions are the following:
\rev{
\begin{itemize}
\item We propose an attack that successfully targets advanced forensic detectors in a black-box scenario, meaning that it is not designed to deceive a specific detector, but it works generally to conceal the tampering traces. 
\item The proposed attack does not require any training phase and does not depend on the access to a pristine dataset for calibration. 
\item The attacker can rely on the single original complex image to achieve outstanding anti-forensic performance.
\item The manipulated images that undergo our proposed attack maintain similar characteristics than pristine data for what concerns their visual quality, \gls{sar} properties and frequency behaviour, thus rendering the attack difficult to be spotted.
\item We validate the proposed pipeline on a dataset of automatically spliced \gls{sar} images, but we also experiment on hand-made forgeries that resemble realistic and potentially alarming scenarios.
\item The proposed attack outperforms another counter-forensic solution lately proposed to conceal local manipulation traces~\cite{tailanian2024diffusion}.
\end{itemize}
}

\vspace{-10pt}
\section{Background}
\label{sec:background}

In this section, we present some background concepts useful to understand the tackled objective and the proposed methodology. 
We start reporting details on \gls{sar} imaging;
then, we show the most common \gls{sar} image manipulations; finally, we describe the existing state-of-the-art forensic detectors that deal with such manipulations. 


\vspace{-5pt}
\subsection{SAR Imaging}
\label{subsec:sar_back}

A \gls{sar} system involves an imaging radar installed on a mobile platform, such as a satellite or an aircraft, moving in a specific direction. As it moves, the system transmits a series of high-power electromagnetic waves through its antenna. These waves engage with the objects they encounter on the Earth's surface and undergo backscattering, altering their amplitude and phase based on the permittivity and physical characteristics of the objects, like geometry and roughness. Subsequently, the antenna captures these modified echoes, which can be processed to generate the complex \gls{sar} image~\cite{bamler1993sar}. 

This final image can undergo additional processing. 
According to the specific acquisition mode and type of additional processing applied, various distinct \gls{sar} products can be generated~\cite{oliver2004understanding}. 
For instance, \gls{sar} images can be acquired with Strip, Spot and Scan acquisition modes~\cite{iceye_modes}, which results in images with different spatial resolution and spectral features. 
Then, \gls{sar} images can be processed to obtain \gls{slc}, \gls{mlc} or \gls{grd} products. 
Among them, \gls{slc} processing considers a single acquisition step per scene, retaining full resolution in azimuth and range directions. \gls{slc} images have the highest fidelity of all \gls{sar} image products, as they are free from interpolation or projection artifacts. 
For this reason, \gls{slc} products are usually employed for \gls{sar} quality assessment, calibration and interferometric applications~\cite{iceye_slc}. 



In the next lines, we first describe the general end-to-end \gls{sar} system model, which refers to the entire pipeline of data acquisition and formation. 
Then, we provide some details on speckle noise, which characterizes all \gls{sar} imaging systems.


\subsubsection{End-to-end SAR system model}
\label{subsubsec:e2esar_model}

The end-to-end \gls{sar} system model regards the modeling of the entire process which starts from the reflection from a scattering object until its conversion into an image. 
The process starts from the source complex scene reflectivity, which collects all the information regarding the object's characteristics (like permittivity, conductivity and surface structure) and the antenna system (e.g., the wavelength of transmitted waves, the polarization, the incidence angle and the imaging geometry)~\cite{bamler1993sar}.
Focusing on \gls{sar} 2D imagery, we can model a generic scene reflectivity image as $\Irc$, where $\Ir \in \mathbb{R}^{X \times Y}$ is the amplitude component and $\mathbf{\Phi}_{R} \in \mathbb{R}^{X \times Y}$ is the phase term. $X$ and $Y$ correspond to range (i.e., the direction perpendicular to platform flight along which the electromagnetic beam travels) and azimuth (the actual trajectory of the platform), respectively.

\begin{figure}[t]
  \centering  \includegraphics[width=\columnwidth]{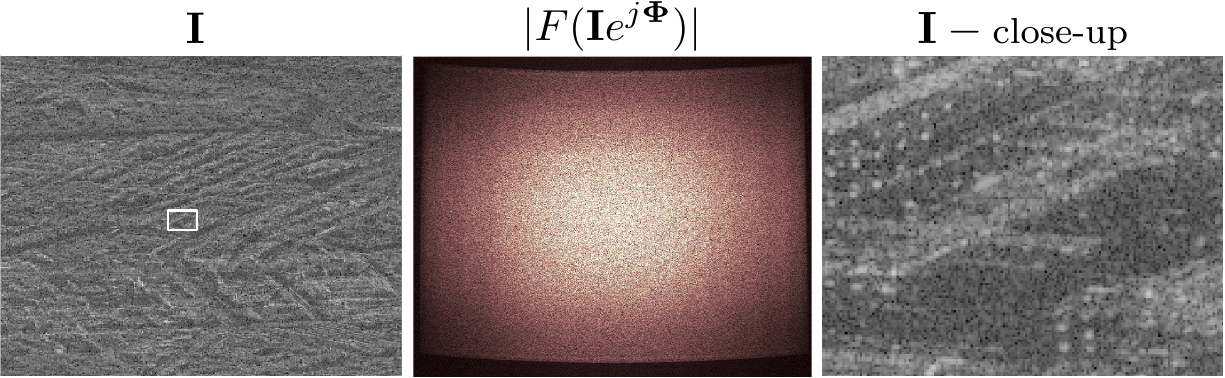}
  \caption{Example of 2D SLC SAR image $\Ic$. First, we report the amplitude component $\I$; then, the amplitude of the \gls{ft} of the complex image $F(\Ic)$; finally, the close-up of $\I$ in the pixel area surrounded by the white box.}
  \label{fig:backg_example}
\vspace{-10pt}
\end{figure}

The scene reflectivity is acquired by the \gls{sar} system in the form of backscattered echoes, which are usually denoted as raw data. The process that regresses raw data back to an estimate of the complex reflectivity scene is named focusing, or image formation. 
In the \gls{sar} imaging community, it is common to model the entire acquisition and processing chain as a cascade of linear filters~\cite{bamler1993sar, guccione2013azimuth, lapini2013blind}. 
The final image spectrum can be related to the initial scene reflectivity spectrum as
\begin{equation}
    F(\Ic) = F(\Irc) \odot \FH(f_x, f_y) , 
    \label{eq:e2esar_model}
\end{equation}
where the operator $F( \cdot )$ describes the 2D \gls{ft} and $\odot$ describes the element-wise product, while $\FH(f_x, f_y)$ is the 2D transfer function of the end-to-end \gls{sar} system model, expressed as a function of the spatial frequencies $f_x$ and $f_y$. The final 2D image $\Ic \in \mathbb{C}^{X \times Y}$ is the result of the \gls{sar} acquisition and focusing operations. An example of the amplitude of a 2D \gls{slc} \gls{sar} image selected from the TerraSAR-X ESA archive~\cite{terrasarx_archive} is reported in Fig.~\ref{fig:backg_example} (left).

In particular, for spaceborne strip-map data $\FH(f_x, f_y)$ can be described as a linear azimuth and range invariant operator, which is usually approximated as a non-negative real central-symmetric 2D filter with a low pass behaviour~\cite{bamler1993sar, lapini2013blind}. 
Formally, 
\begin{equation}
\begin{matrix*}[l]
     \FH(f_x, f_y) \geq 0; \\
     \FH(-f_x, -f_y) = \FH(f_x, f_y); \\
     \FH(f_x, f_y) \approx 0,  \, \, \, \,  \forall |f_x| > f_{c_x}, \forall |f_y| > f_{c_y},
\end{matrix*}
\end{equation}
being $f_{c_x}, f_{c_y}$ specific cutoff frequencies along azimuth and range dimensions, respectively. These frequencies may be known in advance if the \gls{sar} system model is known, but they can be easily inferred by observing the spectrum of the final image $\Ic$~\cite{lapini2013blind}.
Fig.~\ref{fig:backg_example} (center) reports the magnitude spectrum of the previously referred \gls{sar} image. 
It is noticeable the low pass behaviour along both frequency dimensions, though the spectrum is very wide; moreover, in correspondence of horizontal and vertical cutoff frequencies, the magnitude shows a steep drop towards the zero.

\subsubsection{Speckle pattern}
\label{subsubsec:back_speckle}

In the \gls{sar} imaging community, it is well-known that the majority of surfaces exhibit significant roughness at the scale of an optical wavelength~\cite{goodman1976speckle}. 
When exposed to coherent radiation, the wave reflected from such surfaces comprises contributions from numerous independent scattering areas with various different relative delays. The inference of these contributions gives rise to a typical granular pattern known as speckle~\cite{goodman1976speckle}.
This pattern typically presents a continuum of irradiance values, which range from dark spots (due to destructive interference) to bright spots (due to constructive interference). 

The statistical modeling of the speckle pattern is done by considering the sum of the various complex microscopic contributions. Each component carries an amplitude and a phase that are statistically independent of each other and of the amplitude and phase of the other components \cite{goodman1976speckle}. 
Given a large number of contributions, it can be proved that the speckle pattern components (the real and the imaginary part) are independent and identically distributed as a Gaussian probability density function with zero mean. This results in a complex speckle pattern $\Sn$ with an amplitude $\Sa(x, y)$ characterized by a Rayleigh probability density function and a phase term $\mathbf{\Phi}_{S}(x, y)$ uniformly distributed~\cite{dainty1970some}. More specifically, $\Sa(x, y) \sim \mathcal{R}(\sigma_S)$, being $\sigma_S$ the scale parameter of a Rayleigh probability distribution, whether $\mathbf{\Phi}_{S}(x, y) \sim \mathcal{U}[0, 2\pi]$.


If we restrict the field to \gls{slc} \gls{sar} images (i.e., when a single image of the scene reflectivity is acquired and processed by the \gls{sar} system), the scene reflectivity $\Irc$ (i.e., the source data that enters the \gls{sar} imaging system) follows a multiplicative speckle model with a fully developed speckle~\cite{long2015microwave}.
In practice, $\Irc$ results from the element-wise product of a hypothetically noise-free image $\Iideal$ (which can be seen as a non-observable signal in our model), and the speckle pattern. 
Thus, $\Irc = \Iideal \, \odot \, \Sn$.

As a consequence, the scene reflectivity image is composed of the contributions of manifold complex components, each due to a different scattering point on the target surface, showing a speckle pattern superimposed to the image of interest~\cite{goodman1976speckle}. 
This pattern is maintained throughout the process of \gls{sar} data acquisition and focusing, thus it is reflected in the final reconstructed image $\Ic$.
In Fig.~\ref{fig:backg_example} (right) we show a close-up of the previously shown 2D \gls{sar} amplitude image, which is actually an \gls{slc} image, \rev{thus it contains a reasonable amount of speckle noise}. 
\rev{The effect of the speckle pattern is noticeable as} a granular noise superimposed to the scene of interest. 


\vspace{-5pt}
\subsection{SAR amplitude image manipulations}
\label{subsec:manip_back}

\begin{figure}[t]
  \centering  \includegraphics[width=\columnwidth]{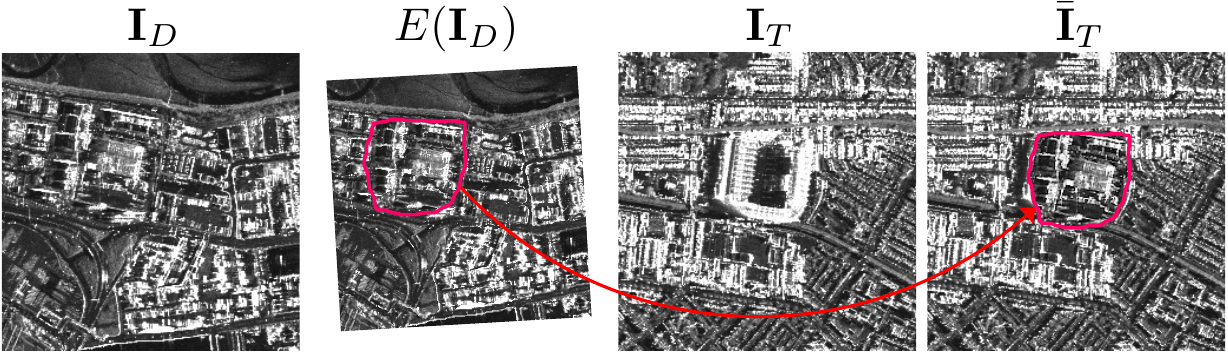}
  \caption{Example of forgery creation process. From left to right: the donor image $\I_D$; the edited version of the donor image $E(\I_D)$; the target image $\I_T$; the locally manipulated target image $\Ibar_T$. The splicing is performed by copying a region of the edited donor image into the target one. The manipulation area is surrounded by the red contour. }
  \label{fig:backg_forgery}
\vspace{-10pt}
\end{figure}


In this paper we deal with the detection of local manipulations on \gls{slc} \gls{sar} amplitude images. 
These images may be manipulated for a variety of reasons, which span from legitimate privacy concerns up to malicious intents guided by political reasons or military purposes in war scenarios.  
As other kinds of digital imagery, these data
can be easily manipulated through editing software suites like Photoshop or GIMP, as well as through synthetic generation tools~\cite{Abady2020, Zhao2021}.

Inspired by~\cite{cannas2022amplitude}, we can formally describe the forgery creation process of a \gls{sar} amplitude image as follows.
We define a generic pristine amplitude \gls{sar} image as $\I$, with size ${X \times Y}$.
Let $\I_D$ and $\I_T$ be two pristine \gls{sar} amplitude images; $\I_D$ is the donor image, whereas $\I_T$ is the target one.
We define $\Ibar_T$ as the tampered with version of $\I_T$, which 
has been locally manipulated by splicing a specific pixel area of $\I_D$ into $\I_T$. 

As done in~\cite{cannas2022amplitude}, we acknowledge the possibility that, before the splicing, the pixels of the donor image might have undergone additional editing procedures like resizing, rotation or noise addition to enhance the credibility and visual appeal of the attack.
For example, rotation and resizing could be necessary to align the content of the source and target images and prevent the splicing from being easily detectable upon visual inspection.
Therefore, we define the generic edited version of the donor image as $E(\I_D)$, with $E(\cdot)$ being a suitable editing function. 
We define the donor pixel region of $E(\I_D)$ as $\DR$ and the target pixel region of $\I_T$ as $\TR$. $\TR$ is the region of $\I_T$ under splicing attack. $\DR$ and $\TR$ are congruent, i.e., the two pixels' regions have the same shape, size and orientation. They can differ in their location inside the respective images $E(\I_D)$ and $\I_T$. 
Fig.~\ref{fig:backg_forgery} provides an example of splicing operation. The original target amplitude image $\I_T$ depicts a strong scatterer in the middle area. This region can be hidden by splicing a portion of the donor amplitude image $\I_D$ into $\I_T$, after the application of little downscaling and rotation to match the dimensions. 

We formally define the resulting spliced image $\Ibar_T$ as:
\begin{equation}
    \label{eq:splicing_def}
    \Ibar_T(x, y) = 
    \begin{cases}
    \DR(x', y') &\quad \text{if $(x, y)\in \TR$}\\
    \I_T(x, y), &\quad \text{if $(x, y)\notin \TR$}
    \end{cases},
\end{equation}
being $(x', y')$ the point coordinates of the donor region corresponding to the target coordinates $(x, y)$.

The splicing operation can be described by a tampering mask $\M$, with the same size of $\I_T$, where each pixel takes a binary value 0 or 1 depending on the pixel of $\Ibar_T$ being pristine or manipulated, respectively.
Formally, the tampering mask $\M$ has pixel values equal to
\begin{equation}
    \M(x, y) = \begin{cases}
    1,\quad \text{if $(x, y)\in \TR$}\\
    0,\quad \text{if $(x, y)\notin \TR$}
    \end{cases}.
    \label{eq:mask_def}
\end{equation}

\vspace{-10pt}
\subsection{SAR forensic detectors}
\label{subsec:detector_back}


The forensic community has developed manifold image splicing detectors during the years, specifically focused on classic $8$-bit imagery like photographs or video frames~\cite{Cozzolino2015, Cozzolino2020, guillaro2023trufor}. 
All the deployed detectors leverage the same assumption, that is, the acquisition device and processing operations leave peculiar traces on images, and these can be exploited to expose local forgeries~\cite{Lukas2006, Cozzolino2020}. 

The acquisition pipeline of \gls{sar} amplitude images is completely different than digital photographs or video sequences. 
Nonetheless, it has been shown that different \gls{sar} products reasonably contain different traces relative to the processing executed for generating them~\cite{cannas2022amplitude}. 
In this vein, the authors of~\cite{cannas2022amplitude} recently proposed \rev{two forensic detectors, namely the \gls{sae} and the \gls{asae}}, 
for localizing splicing areas on \gls{sar} amplitude images. 
\rev{Following a similar approach, the TruFor detector~\cite{guillaro2023trufor},
which has shown outstanding performances on natural images and represents now the state of the art on them~\cite{edwards2024review},
can be adapted to \gls{sar} amplitude images, provided that the training pipeline is specifically designed to account for the unique characteristics of \gls{sar} data. We refer to this \gls{sar}-adapted version of the detector as \gls{truforsar}\footnote{More details are reported in the supplemental materials.\label{footnote_truforsar}}}.

\rev{These detectors highlight} local inconsistencies between the donor and target pixel areas. 
More formally, let us consider a generic manipulated amplitude image as input to the detector.
For simplicity, we omit the subscript $T$ from the manipulated images notation, taking for granted that all amplitude images indicated as $\Ibar$ have been manipulated in a specific target region of pixels, identified by the tampering mask $\M$. 
\rev{The three detectors estimate a real-valued heatmap $\Pbar$, ranging in $[0, 1]$, 
that identifies the forgery region}. 
\rev{Examples of the tampering heatmap $\Pbar$ estimated by the \gls{truforsar} detector can be found in Figs.~\ref{fig:intro_example} and~\ref{fig:problem_example}.}

\section{Problem Formulation}
\label{sec:problem}

\begin{figure}[t]
  \centering  \includegraphics[width=.6\columnwidth]{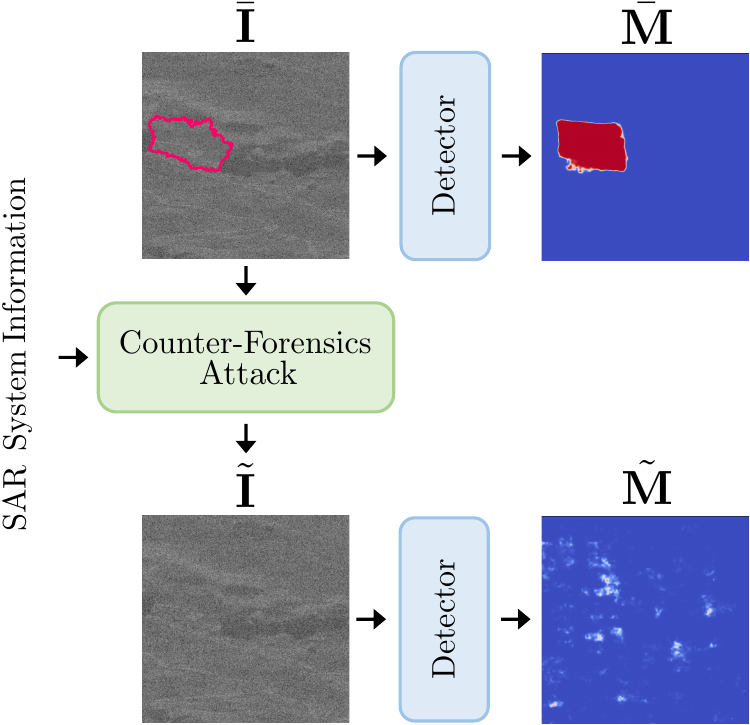}
  \caption{Sketch of the tackled counter-forensic task. Given a manipulated amplitude image $\Ibar$ and some information on the SAR system used to acquire the original data, we aim at producing an amplitude image $\Itilde$ in which the manipulation traces have been concealed. 
}
  \label{fig:problem_example}
\vspace{-10pt}
\end{figure}

In this paper we tackle a counter-forensic task, aiming at hindering the capability of existing forensic detectors in spotting local forgeries applied to \gls{sar} amplitude images.
We assume to have available an amplitude image $\Ibar$, which has been locally manipulated for inserting or hiding details, as shown in Section~\ref{subsec:manip_back}. Moreover, we suppose to know which is the satellite that has been used to acquire the original data and which are the acquisition mode (e.g., strip, spot, scan) and processing applied (e.g., \gls{slc}, \gls{mlc}, or \gls{grd}).
Notice that this situation resembles a realistic scenario, as the potential attacker typically has knowledge of the \gls{sar} system used to acquire the data under manipulation.

Our goal is deleting the manipulation artifacts from $\Ibar$ such that forensic detectors cannot spot the manipulation traces anymore. 
In doing so, we do not want to deviate from neither the semantic content nor the visual quality of $\Ibar$. Moreover, we want to transfer to the attacked image the same properties of the pristine data acquired by the original \gls{sar} system, like the speckle noise and the traces left by the acquisition and processing operations.  
In other words, the final produced amplitude image must completely match the semantic characteristics of $\Ibar$ (i.e., it actually presents a forged area), but it has to be transparent to visual inspection (i.e., the visual quality is maintained), to \gls{sar} properties inspection, and to the forensic detector, that must estimate it being pristine in its entire area. We define the attacked amplitude image as $\Itilde$.

We consider as reference \rev{detectors those listed in}
Section~\ref{subsec:detector_back}, because \rev{they are}
specially tailored to \gls{sar} imagery and represent the state-of-the-art solutions for spotting local image manipulations in the \gls{sar} field.
\rev{Our goal turns into concealing the forensic traces from $\Ibar$ such that its estimated heatmap resembles that of a pristine image}. 

Fig.~\ref{fig:problem_example} reports the sketch of the tackled task: given a manipulated amplitude image $\Ibar$, the forensic \rev{detectors return a real valued heatmap $\Pbar$} that highlights some tampering traces in correspondence of the manipulated pixels. 
Our proposed method leverages information on the \gls{sar} system used to acquire the original data for performing a counter-forensic attack on $\Ibar$. 
The result is the amplitude image $\Itilde$ which visually looks the same as $\Ibar$ and has the same \gls{sar} properties of a pristine image; however, if passed from the \rev{detectors, the heatmap $\Ptilde$} does not show any manipulation artifacts. 
We provide more details on the proposed counter-forensic attack in the next section.

\section{Proposed Methodology}
\label{sec:method}
\begin{figure}[t]
  \centering  \includegraphics[width=\columnwidth]{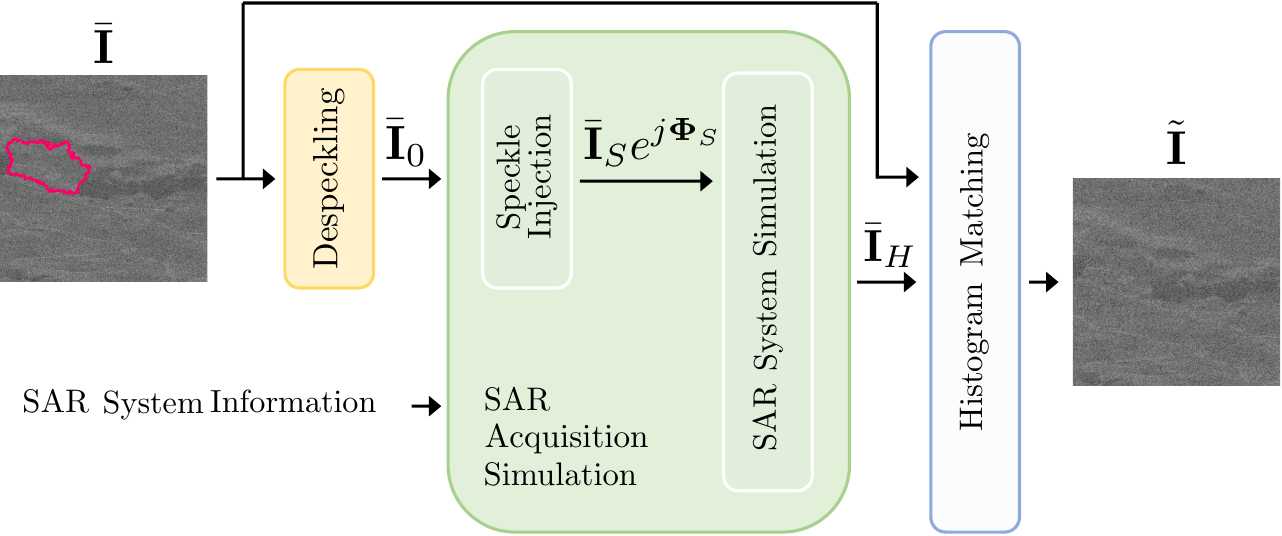}
  \caption{Scheme of the proposed counter-forensic attack. By exploiting information of the original SAR system, we recreate the effect of a SAR acquisition through the same SAR system (or an estimate of it) used to acquire pristine data. To do so, we apply despeckling to the input manipulated image, then we simulate a re-acquisition by injecting speckle noise and filtering with the end-to-end SAR system. Finally, we match the histogram of the produced amplitude image with that of the manipulated amplitude image $\Ibar$. 
}
  \label{fig:method_pipeline}
 \vspace{-10pt}
\end{figure}
To hide local manipulations from a \gls{sar} amplitude image $\Ibar$, we propose to recreate the effect of a \gls{sar} acquisition of the scene targets depicted in $\Ibar$. 
To do so, we exploit the original \gls{sar} system information which  is available at attacker side. 
The output of the \gls{sar} acquisition is the amplitude image $\Itilde$, 
which still resembles $\Ibar$ from the semantic content viewpoint (both in tampered with and original areas) but presents \gls{sar}-related properties that are equivalent to those of pristine data acquired with that satellite.



A sketch of the proposed methodology is shown in Fig.~\ref{fig:method_pipeline}. Our method is composed of three main steps:
\begin{enumerate}
    \item \textit{Image despeckling}. We remove the speckle noise from the manipulated amplitude image $\Ibar$ to recreate the noise-free scene reflectivity image as it was before acquisition.
    \item  \textit{SAR acquisition simulation}. By leveraging information on the original \gls{sar} system, we simulate the acquisition of the targets depicted in the manipulated scene. 
    \item \textit{Histogram matching}. We match the pixel histogram of the resulting image with that of the manipulated amplitude image $\Ibar$, in order to keep the dynamic range similar to the initial one. This step can be seen as a refinement phase of the previous point.
\end{enumerate}

The effect of the proposed methodology is twofold.
On the one hand, by acquiring the scene targets depicted in $\Ibar$ with the same acquisition system of the pristine data, we force the resulting amplitude image $\Itilde$ to maintain similar \gls{sar} characteristics to the pristine data, like the amount of speckle noise and its spatial and frequency pattern. The final histogram matching completes the process by adjusting the pixel dynamics.
On the other hand, the \gls{sar} acquisition processing chain acts as a counter-forensic attack that, by slightly modifying the pixel low-level information which are paramount for the \rev{forensic detectors}, enables to hide the tampering traces. 
A detailed explanation of all the steps follows in the next lines. 


\vspace{-5pt}
\subsection{Image Despeckling}
\label{subsec:method_despeckling}
Since the goal is reproducing the effect of a \gls{sar} acquisition of the scene targets depicted in the manipulated image, the first natural step of an attacker would be to remove the speckle noise introduced by the previous acquisition.
This operation would enable to simulate the acquisition of noise-free targets, which is exactly what happens if we consider the \gls{sar} acquisition model presented before. 

To do so, we leverage well-known state-of-the-art \gls{sar} despeckling algorithms,
namely the \gls{fans}~\cite{cozzolino2013fast} and the \gls{nlsar}~\cite{deledalle2014nl}. 
Both techniques reported excellent performances on 
\gls{sar} data and can work with the image amplitude information alone. These characteristics make them perfectly suited for our case study with respect to modern data-driven solutions~\cite{Molini2022} that would require a 
specific training 
and
work with complex images exclusively.
Following a similar notation to Section~\ref{subsubsec:back_speckle}, the despeckled amplitude image is defined as $\Ibard$.

Notice that the main disadvantage of image despeckling is the potential loss of high frequency details in the final despeckled image. 
As a matter of fact, the despeckling algorithm risks to remove also point scatterers that were originally depicted in the reflectivity scene. 
We show in our experiments that this despeckling step can be avoided, as the effects of the previous \gls{sar} acquisition are completely superseded by the new acquisition step that follows.
To keep at most the frequency content of the scene depicted in the manipulated image $\Ibar$, we prove that it is a better choice to omit this step, directly going to the simulation of the \gls{sar} acquisition. 

\vspace{-5pt}
\subsection{SAR Acquisition Simulation}
This step is the main core of the proposed counter-forensic attack. It consists of recreating the acquisition of the scene targets depicted in the manipulated image by means of the same \gls{sar} system (or an estimate of it) that was used to acquire the original data.
Referring to \gls{sar} background concepts reported in Section~\ref{subsec:sar_back}, we split this step into two separate stages, namely speckle injection and \gls{sar} system simulation.

\subsubsection{Speckle Injection}
\label{subsubsec:method_speckle_inj}
The injection of speckle noise is required to simulate the acquisition of a complex \gls{sar} image in which the targets, defined by the noise-free manipulated amplitude image $\Ibard$, are affected by speckle noise.
To do so, we generate a complex speckle noise $\Sn$ with the same size of $\Ibard$ and with the characteristics reported in Section~\ref{subsec:sar_back}.
In general, $\Sa(x, y) \sim \mathcal{R}(\sigma_S)$, being $\sigma_S$ the scale parameter of a Rayleigh distribution, whether $\Phi_{\mathbf{S}}(x, y) \sim \mathcal{U}[0, 2\pi]$.

The output image of this stage is obtained by element-wise multiplication between the input image $\Ibard$ and the speckle noise. This is defined as $\Ibarsc = \Ibard \odot \Sn$, being $\odot$ the element-wise product.
Notice that the final image $\Ibarsc$ is complex, i.e., it carries both amplitude and phase information. 
Moreover, the complex image returned by this step has an extremely wide spectrum, due to the multiplication by a complex term with uniformly distributed phase in all frequencies. 



\subsubsection{SAR System Simulation}
\label{subsubsec:method_sar_simulation}
To recreate a \gls{sar} acquisition, the speckle-affected image $\Ibarsc$ must be filtered through the same acquisition and processing system used to acquire the original data. 
This maps into the exploitation of the \gls{sar} system frequency response $\FH$ described in Section~\ref{subsubsec:e2esar_model}.

\begin{figure}[t]
  \centering  \includegraphics[width=\columnwidth]{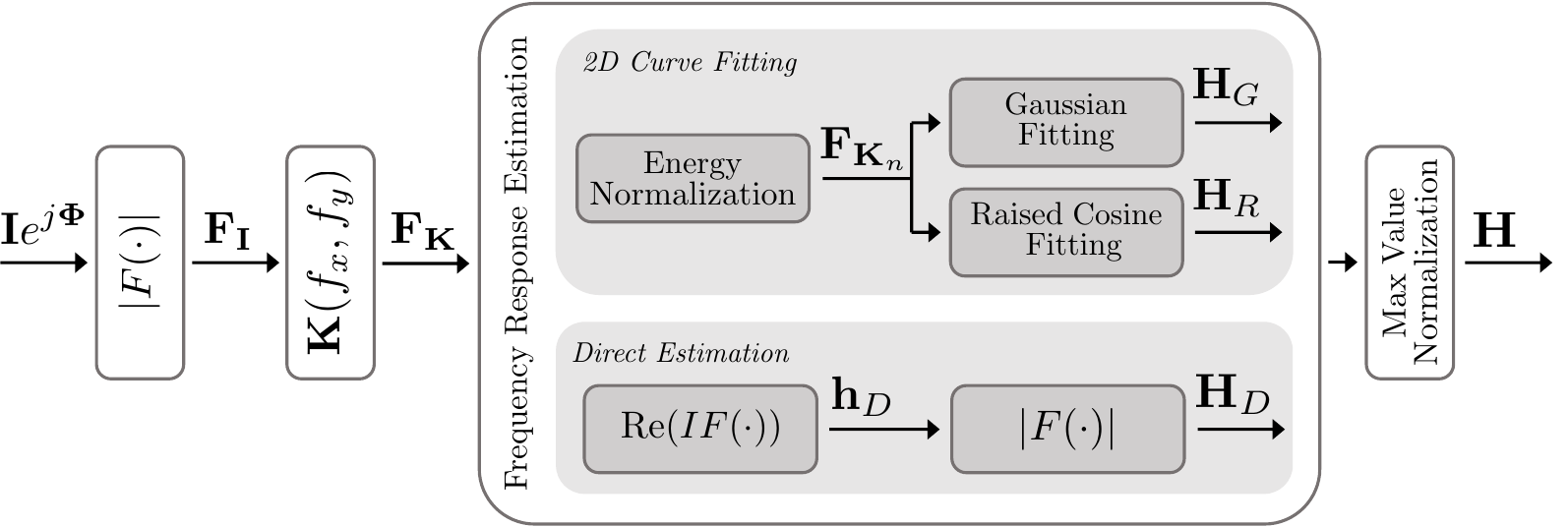}
  \caption{Proposed pipeline for estimating the frequency response of the \gls{sar} system used to acquire $\Ic$. The final estimated acquisition system is defined as $\FH$. 
  }
  \label{fig:method_sar_acquisition}
  \vspace{-10pt}
\end{figure}

In this context, we consider four different realistic scenarios that the attacker can encounter: (i) $\FH$ is fully known; (ii) the attacker disposes of the original \textit{complex} image under attack; (iii) the attacker disposes of $N$ complex images captured by the same satellite, different from the original image under attack; (iv) the attacker only disposes of the original \textit{amplitude} image under attack. 
The last three scenarios require an estimation process for extracting information on the original system frequency response. 
As suggested in \cite{lapini2013blind}, we assume the \gls{sar} system frequency response can be approximated by a real central–symmetric nonnegative function. 
For simplicity, we describe the proposed estimation process for case (ii), detailing the differences with respect to cases (iii) and (iv) at the end of the derivations. 

To estimate $\FH$, we consider four consecutive steps (see Fig.~\ref{fig:method_sar_acquisition}):
\begin{enumerate}
    \item We first compute the 2D \gls{ft} of the original image $\Ic$ and take its magnitude value, defined as $\F_{\I} = |F(\Ic)|$. 
    \item We apply a smoothing to $\F_{\I}$ through a Gaussian kernel $\mathbf{K}$. Formally, we can define the resulting magnitude spectrum as 
    \begin{equation}
        \F_{\mathbf{K}}(f_x, f_y) = \F_{\I}(f_x, f_y) \ast \mathbf{K}(f_x, f_y), 
    \end{equation}
    being $\ast$ the 2D linear convolution and $(f_x, f_y)$ the tuple of spatial frequencies.
    We do so for reducing the local variability of the acquired data. This helps untying the filter estimation process from the actual data content which is noisy and dependent on the depicted scene.
    \item Given $\F_{\mathbf{K}}$, we estimate three different versions of the frequency response of the \gls{sar} acquisition system: (i) we fit a 2D Gaussian bell; (ii) we fit a 2D Raised Cosine, as suggested in \cite{lapini2013blind}; (iii) we directly estimate it by slightly modifying $\F_{\mathbf{K}}$.

\textbf{2D Curve Fitting.}
Gaussian and Raised Cosine fitting are done through the same sequence of operations. 
In particular, we need
two further steps:
\begin{itemize}
    \item As done in~\cite{lapini2013blind}, we normalize $\F_{\mathbf{K}}$ by the square root of its energy, obtaining 
    \begin{equation}
    \F_{\mathbf{K}_n} = \frac{\F_{\mathbf{K}}}{\sqrt{\sum_{f_x} \sum_{f_y} \F^2_{\mathbf{K}}(f_x, f_y)}}.    \end{equation}
    \item Through least squares fitting, we exploit $\F_{\mathbf{K}_n}$ to fit a 2D Gaussian bell $\mathbf{G}(f_x, f_y)$ or a 2D raised cosine $\mathbf{R}(f_x, f_y)$. Both functions present separable components along the two frequency dimensions. $\mathbf{G}(f_z)$ is defined as
    \begin{equation}
    \mathbf{G}(f_z) = g_z \cdot e^{- \frac{(f_z - \mu_z)^2}{2 \sigma^2_z}}, 
    \end{equation}
    where $z \in \{x, y \}$ and being $g_z, \mu_z$ and $\sigma_z$ the gain, the mean value and the variance along $z$ dimension, respectively.
    $\mathbf{R}(f_z)$ is defined as
    \begin{equation}
    \mathbf{R}(f_z) = (A_z - B_z \cos(\pi ( f_z - f_{c_z})/f_{c_z})), \,  f_z \leq f_{c_z},
    \label{eq:raised_cosine}
    \end{equation}
    where $f_{c_z}$ is the cutoff frequency ($\mathbf{R}(f_z) = 0$ for $fz > f_{c_z}$) and $A_z > 0 $ and $B_z > 0$ are model parameters \cite{lapini2013blind}. 
\end{itemize} 
We define the 2D frequency responses achieved by the least squares fitting as $\FHg$ and $\FHr$, considering Gaussian and Raised Cosine, respectively. These frequency responses always correspond to real-valued central-symmetric nonnegative functions.

\textbf{Direct estimation.}
In the last case, we directly employ $\F_{\mathbf{K}}$ to estimate the frequency response of the \gls{sar} acquisition filter. 
Notice that, being $\F_{\mathbf{K}}$ the frequency response of a \textit{complex} signal, it does not necessarily correspond to a central-symmetric function. 
To circumvent this issue, we compute its 2D \gls{ift} and consider only the real part contribution. Formally, we define the 2D real signal 
\begin{equation}
    \h_D = \mathrm{Re} (IF(\F_{\mathbf{K}})),
\label{eq:direct_estimation}
\end{equation}
where $IF( \cdot) $ represents the 2D Inverse Fourier transform operator. 
The finally estimated 2D frequency response is defined as $\FHd = |F(\h_D)|$. In this case, $\FHd$ satisfies the symmetry and positivity constraints. 

\item In the last phase,
the \gls{sar} acquisition filter is estimated by normalizing the obtained frequency response, i.e., either $\FHg$, $\FHr$ or $\FHd$, by its maximum value, such that all versions have maximum gain equal to $1$.   
We denote the estimated system frequency response as $\FH$. 
In the Gaussian case, $\FH = \FHg / \FHgmax$; in the Raised Cosine case, $\FH = \FHr / \FHrmax$; in the direct estimation case, $\FH = \FHd / \FHdmax$.


\end{enumerate}

If the original complex image is not available at attacker side but $N$ complex images captured by the same satellite can be found, $\FH$ can be estimated as the arithmetic mean of the single frequency responses estimated by each complex image. 

In the case in which only the original amplitude image is known by the attacker, the estimation process is strongly affected by the loss of high frequency details in the spectrum. We show in Section~\ref{subsec:results_quality} that the amplitude image is not sufficiently informative for estimating the \gls{sar} system, thus returning poor results in the simulation of the re-acquisition.



\vspace{-10pt}
\subsection{Histogram Matching}
In the last phase of the proposed attack, it may happen that the resulting amplitude image $\Ibarh$ presents a few differences with respect to the image $\Ibar$ in terms of dynamic range and pixels' distribution. 
As a matter of fact, there may be little differences between the injected and the actual speckle noise distributions. Also, the estimation of the system frequency response may contain small errors.  
All these situations contribute to little mismatch between the distribution of attacked and non-attacked amplitude images, thus resulting in clear artifacts that may induce forensic investigators to expose the attacked ones. 

To counteract this effect, we match the pixels' histogram of $\Ibarh$ with that of the manipulated amplitude image $\Ibar$.
In doing so, the dynamics and the pixel distribution are adapted to the initial ones, hindering the traces of the global operation done to hide the forgery. 

We define the final amplitude image as $\Itilde$. 
This image still presents the forgered area, though, as we show in our experiments, the splicing area is invisible to the existing forensic detectors developed for image splicing localization.


\section{Experimental Setup}
\label{sec:setup}

In this section, we describe the setup of our experimental campaign. First, we present the employed \gls{sar} image dataset; second, we provide details on the forgery creation process; third, we give information on the counter-forensic attack parameters; finally, we describe the forensic \rev{detectors} setup. 

\vspace{-10pt}
\subsection{Pristine Dataset}

We downloaded the data from ESA EO-CAT \cite{eocat_esa}, which provides various \gls{sar} products acquired in different modes and missions.
In particular, we selected the \gls{sar} \gls{st} \gls{ssc} products from the TerraSAR-X ESA archive~\cite{terrasarx_archive}; these products present a resolution of $0.25$m and a $16$-bit dynamic range.  
We purposely selected these products for three main reasons: 
\begin{itemize}
    \item They are acquired with the highest possible spatial resolution, enabling us to produce forgeries in the area of few meters.
    \item They are single look, meaning that the resulting \gls{sar} images are not the average of multiple satellite looks and thus present a reasonable amount of speckle noise.
    \item They are complex products, carrying both amplitude and phase information. 
\end{itemize}

Overall, there are $24$ different ST-SSC products available. The temporal coverage ranges from $2016$ to $2021$ and the scene content varies between city center, seaside, cropland and mountain. In particular, these products show images of four specific geographical areas, located in Rome and Dublin cities, in Gaza and in north-Italy mountains. 

The size of ST-SSC products is not fixed and can be larger than $10K \times 18K$ pixels. Given the huge size of these acquisitions, we divided each of them into $1024\times 1024$ pixels-wide images with an overlap of $512\times 512$ pixels. 
This operation enables to operate at a local level with fine granularity, facilitating easy processing of the input by our networks.
From each product, we extracted $400-500$ complex images, resulting in a total set of approximately $12K$ pristine data.
Some examples are shown in the first column of Fig.~\ref{fig:setup_landcover_forgeries}~\footnote{More examples are reported in the supplemental materials.\label{footnote_supp_examples}}.

\vspace{-5pt}
\subsection{Manipulated Dataset}
\label{subsec:setup_manipulated_dataset}
To create realistic forgeries, we always considered the case in which the donor amplitude image $\I_D$ and the target amplitude image $\I_T$ come from the same product. 
This operation allows to produce realistic splicings where the difference in the dynamic range and semantic details is reasonably small, since both donor and target images belong to the original product. 
To run multiple experiments, we created automatic forgeries by randomly selecting donor and target regions of $128 \times 128$ pixels, as done in \cite{cannas2022amplitude}. 

As done in \cite{cannas2022amplitude}, we simulated an attacker perspective by processing the donor amplitude images with post-processing operations which can make the tampering more plausible. 
In particular, recalling Section~\ref{subsec:manip_back}, we applied four possible editings $E( \cdot)$ to the donor image: a Gaussian blur, an upscaling, a downscaling and a rotation. 
Each of the last three similarity transformations is proposed in two different versions: (i) a ``Near'' version, in which the similarity parameters limit the transformations to almost imperceptible changes in the edited images, like a few degree rotation; (ii) a ``Far'' version, in which the editing process becomes more noticeable. 

\begin{figure}[t!]
\centering
    \begin{subfigure}[b]{\columnwidth}
        \centering
        \includegraphics[width=\columnwidth]{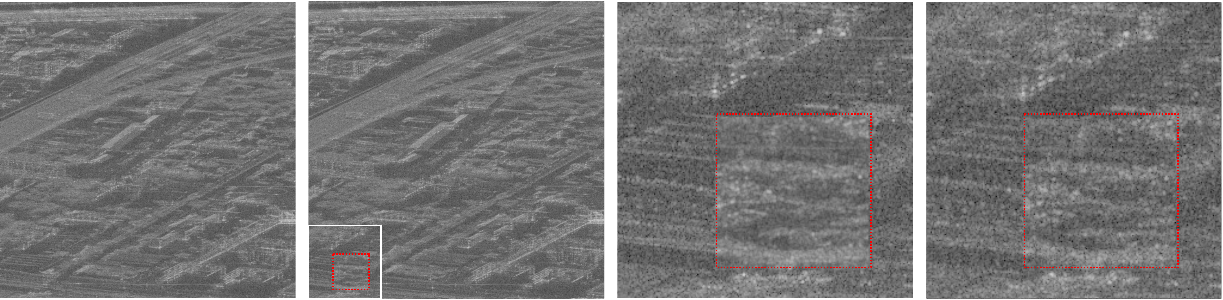}
    \end{subfigure}
\hfill
    \begin{subfigure}[b]{\columnwidth}
        \centering
        \includegraphics[width=\columnwidth]{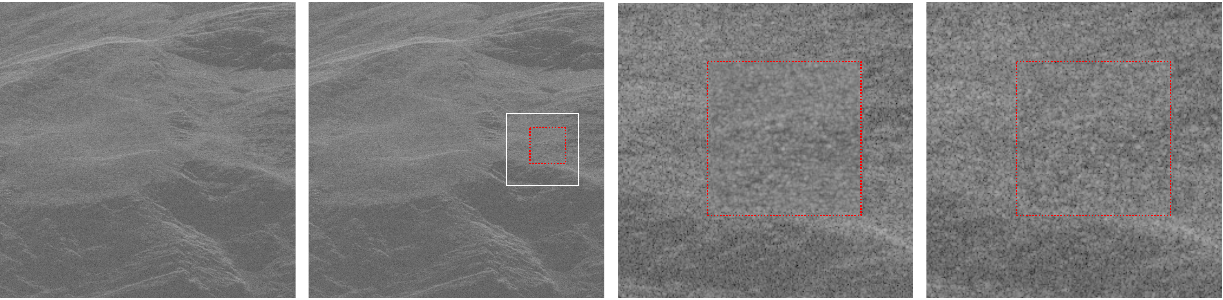}
    \end{subfigure}
\caption{Examples of the considered \gls{sar} amplitude data. First column depicts pristine images.
Second column shows manipulated version of the same images; the splicing area is contained inside the red box.
Third column shows a close-up of the forgery area and of the surrounding pixels (white box). 
Last column depicts a close-up with the proposed counter-forensic attack. 
Since the considered images have $16$-bit dynamics, we show their amplitudes in logarithmic scale.}   
\label{fig:setup_landcover_forgeries}
\vspace{-10pt}
\end{figure}

In any case, all the considered editing operations do not introduce important content variations or degradations in the image visual quality. 
We consider this last constraint being paramount for performing realistic investigations. 
Indeed, we could think to include strong image perturbations like heavy noise addition or blurring. Such editing operations would carry so many artifacts that the manipulation traces could be exposed by any \gls{sar} expert looking at the image. In this scenario, there would be no reason to perform counter-forensic attacks on the manipulated images, as the image content would be so altered that any state-of-the-art detector could spot the forgery.
Table~\ref{tab:setup_editing} reports the parameters used for executing the editing operations. 
Examples of automatic forgeries with gaussian blur editing are shown in second and third column of Fig.~\ref{fig:setup_landcover_forgeries}\footref{footnote_supp_examples}. 

\vspace{-10pt}
\subsection{Counter-forensic attack parameters}
\subsubsection{Image despeckling}
The two despeckling algorithms considered, i.e., \gls{fans} and \gls{nlsar}, are both unsupervised and required no fine-tuning. We set the number of looks to $1$, considering our TerraSAR-X \gls{slc} samples. For the \gls{nlsar}, we had to process each sample as non-overlapping patches of $128\times 128$ pixels for computational reasons. The final despeckled image is obtained by reassembling all the patches together. \gls{fans} instead worked seamlessly on the full resolution images.

\subsubsection{SAR System Frequency Response}
The 2D Gaussian kernel $\mathbf{K}$ used for smoothing the pristine image spectrum has $\sigma = 100$ pixels and size $601 \times 601$ pixels.  
To fit the 2D Gaussian bell and the 2D Raised Cosine, we employed a nonlinear iterative least squares solver, whose parameters are defined in \cite{lsqr_fit}.


\begin{table}[t]
\centering
\caption{Editing operations considered during the process of image local manipulation.}
\label{tab:setup_editing}
\resizebox{.7\columnwidth}{!}{
\begin{tabular}{@{}ll@{}}
\toprule
Editing operation & Parameters \\
\midrule
Gaussian Blur & $\sigma = 0.5$ \\
Upscale Near & Resize factor $\sim \mathcal{U}[1.05, 1.5)$\\
Upscale Far & Resize factor $\sim \mathcal{U}[1.5, 2]$\\
Downscale Near & Resize factor $\sim \mathcal{U}[0.65, 0.95]$\\
Downscale Far & Resize factor $\sim \mathcal{U}[0.5, 0.65)$\\
Rotate Near & Rotation angle $\sim \mathcal{U}[5, 15)$ deg\\
Rotate Far & Rotation angle $\sim \mathcal{U}[15, 45]$ deg\\
\bottomrule
\end{tabular}}
\vspace{-10pt}
\end{table}

\vspace{-15pt}
\rev{\subsection{Detectors' setup}}
In our experiments, we used the detectors in the best possible conditions, i.e., we trained and tested them over a homogeneous dataset, in which the same product was available both in training and testing stages. 
In doing so we guaranteed that, in absence of counter-forensic attacks, the detectors performed at the best of their possibilities. 
Indeed, we wanted to avoid the risk of introducing errors due to different data distributions in train and test stages.
This methodology ensured that the unique factor which might cause a degradation in the performance is the proposed counter-forensic attack. 

To train and test the forensic detectors in the best possible conditions, 
we always showed in training phase all the possible semantic scenarios: for each of the four considered geographical areas (Rome, Dublin, Gaza and Italy mountains), we randomly selected $70\%$ of data for training (further split in $60-40\%$ for training and validation sets, respectively) and left the remaining $30\%$ for testing. 

\rev{Due to their specific training paradigm, all the detectors 
must be trained over locally manipulated images\footref{footnote_truforsar}. To do this, we applied local manipulations over the training and validation sets as explained in~\ref{subsec:setup_manipulated_dataset}. For each amplitude image, we considered one possible splicing area, performing on the donor image all the editing operations listed in Table~\ref{tab:setup_editing}.}

\rev{The images of the test set did always present local manipulations. 
This operation allows to test the detection performances in presence of tampering: if this is not the case (i.e., the test image is pristine), the detectors would report an almost flat heatmap due to absence of inconsistencies~\cite{cannas2022amplitude, guillaro2023trufor}. The tampering operations were realized as explained above.}

\vspace{8pt}
\section{Results}
\label{sec:results}

\subsection{Quality of the attacked images}
\label{subsec:results_quality}



At first, we aim at evaluating the ability of our proposed counter-forensic attack in producing high-quality attacked amplitude images. 
As previously stated, we must produce attacked images which report similar characteristics than original pristine ones, such that an expert in the \gls{sar} field would not be able to easily spot that some global post processing operations have been applied. 

To do so, we put aside for a moment the locally manipulated amplitude images, focusing only on pristine samples. For each pristine amplitude image $\I$, we generate an attacked version of it, namely $\Itilde$, by following the methodology proposed in Section~\ref{sec:method}. 
Therefore, in this particular experiment, the input to the counter-forensic attack are the pristine amplitude images $\I$ and not their manipulated versions $\Ibar$. 
If the attacked versions of pristine samples show indistinguishable from the pristine ones, our attack can be employed to realistically modify \gls{sar} data samples that resemble original untouched ones. 

\textbf{Image quality metrics. }
As done in~\cite{cannas2023deep}, we compare pristine and attacked samples by exploiting several similarity metrics\footnote{In the supplemental materials,  
we also compare attacked and pristine images for what concerns their frequency behaviour.}.
We evaluate the visual quality of the attacked samples by computing metrics 
typical of the image full-reference quality assessment field, 
i.e., 
the \rev{\gls{psnr}} and the \gls{mssim}~\cite{Wang2003multi}. 
\rev{The higher these metrics, the better the quality of the attack, in terms of producing pixel content that visually resembles original data. Notice that the \gls{psnr} has to be intended in the uint16 dynamic range, thus we expect higher values with respect to those typically occurring in counter-forensics for uint8 images~\cite{chervyakov2020analysis, cao2024transferable, fang2023attacking, tailanian2024diffusion}.}

We also consider \gls{sar} compatibility metrics to evaluate our ability in reproducing speckle pattern in the attacked images. In particular, we rely on the \gls{enl}~\cite{Gagnon1997SpeckleFO}, usually employed in the \gls{sar} despeckling field, to understand if the considered attacked samples are compatible with the original \gls{sar} data.
The \gls{enl} is a statistical measure to quantify the level of speckle noise in \gls{sar} images. 
In particular, we compute the absolute relative difference of the \gls{enl} of attacked images with respect to their original pristine version. 
We define this metrics as \rev{$|\Delta \text{-}\textrm{ENL}| = |(\textrm{ENL}_{pristine} - \textrm{ENL}_{attacked}) / \textrm{ENL}_{pristine}|$}.
Our goal is achieving low values of $|\Delta \text{-}\textrm{ENL}|$, meaning for a realistic reproduction of speckle noise in the attacked samples.  




\textbf{Despeckling influence. }
We start evaluating the achieved metrics according to presence or not of the despeckling step in our attack. 
To do so, we restrict our investigations to one of the four situations presented in Section~\ref{subsubsec:method_sar_simulation}, i.e., the different levels of attacker knowledge on the \gls{sar} system frequency response $\FH$.
In particular, we omit the case in which $\FH$ is fully known, as we have no access to all the information describing the original \gls{sar} system used to capture the considered dataset. We consider the best case scenario, which reasonably consists in having available the original complex image $\Ic$ under attack.

\begin{table}[t]
\caption{Quality metrics (\rev{PSNR [dB]} / MSSIM / $|\Delta \text{-}\textrm{ENL}|$) between pristine amplitude images and their attacked versions in presence or absence of despeckling. We hypothesize the attacker has available the original complex image $\Ic$ for estimating the frequency response $\FH$. In bold, the best result per column.}
\label{tab:result_quality_despeckling}
\centering
\resizebox{\columnwidth}{!}{
\begin{tabular}{@{}lccc@{}}
\toprule
                &    \multicolumn{2}{c}{W/ Despeckling}
                & W/O Despeckling 
                \\ \cmidrule(l){2-4} 
              &   \multicolumn{1}{c}{FANS} & \multicolumn{1}{c}{NL-SAR} & \multicolumn{1}{c}{--} \\
              \midrule
              &   \multicolumn{1}{c}{\rev{PSNR} / MSSIM / $|\Delta \text{-}\textrm{ENL}|$} & \multicolumn{1}{c}{\rev{PSNR} / MSSIM / $|\Delta \text{-}\textrm{ENL}|$} & \multicolumn{1}{c}{\rev{PSNR} / MSSIM / $|\Delta \text{-}\textrm{ENL}|$} 
\\ 
\midrule
$\FH_G$   &   $\mathbf{50.90}$ / $\mathbf{0.997}$ / $306.2\%$                 & $\mathbf{50.65}$ / $\mathbf{0.997}$ / $414.5\%$                 & $\mathbf{52.64}$ / $\mathbf{0.998}$ / $0.518\%$        \\
$\FH_R$   & $50.83$ / $0.997$ / $\mathbf{305.2}\%$               & $50.59$ / $0.997$  / $\mathbf{412.7}\%$                 & $52.46$ / $0.998$ / $0.471\%$          \\
$\FH_D$  & $50.87$ / $0.997$  / $305.7\%$              & $50.62$ / $0.997$ / $413.7\%$  & $52.55$ / $0.998$  / $\mathbf{0.458\%}$              \\ \bottomrule
\end{tabular}
}
\vspace{-10pt}
\end{table}

We consider
three different estimation strategies of the \gls{sar} system frequency response, namely the Gaussian fitting, the Raised Cosine fitting and the direct estimation, leading to $\FH_G$, $\FH_R$ and $\FH_D$ respectively.
As regards the speckle noise injection parameters, we show results for phase-only injected speckle, as this scenario achieves far better results than cases with speckle amplitude contributions\footnote{We report a more complete analysis in the supplemental materials.\label{footnote_supp_generic}}.

\begin{table*}[t]
\caption{Quality metrics (\rev{PSNR [dB]} / MSSIM / $|\Delta \text{-}\textrm{ENL}|$) between pristine amplitude images and their attacked versions. In bold, the best results per column. }
\label{tab:result_quality_filter_estimation}
\centering
\resizebox{.9\textwidth}{!}{
\begin{tabular}{@{}lcccccc@{}}
\toprule
                 & &  \multicolumn{4}{c}{Attacker availability for $\FH$ estimation}   \\ \cmidrule(l){3-6} 
             &  & \multicolumn{1}{c}{$\Ic$ (original \textit{complex})} & \multicolumn{1}{c}{$\Ici \ne \Ic$} & \multicolumn{1}{c}{$ \{\Ici \}_{i=1}^N \ne \Ic$ ($N=23$)} & \multicolumn{1}{c}{$\I$ (original \textit{amplitude})} \\ \midrule
          &  & \multicolumn{1}{c}{\rev{PSNR} / MSSIM / $|\Delta \text{-}\textrm{ENL}|$} & \multicolumn{1}{c}{\rev{PSNR} / MSSIM / $|\Delta \text{-}\textrm{ENL}|$} & \multicolumn{1}{c}{\rev{PSNR} / MSSIM / $|\Delta \text{-}\textrm{ENL}|$} & \multicolumn{1}{c}{\rev{PSNR} / MSSIM / $|\Delta \text{-}\textrm{ENL}|$}
\\ 
\midrule
$\FH_G$   &  & $\mathbf{52.64}$ / $\mathbf{0.998}$ / $0.518\%$                 & $\mathbf{52.67}$ / $\mathbf{0.998}$ / $0.506\%$                 & $\mathbf{52.64}$ / $\mathbf{0.998}$ / $0.519\%$         &   -       \\
$\FH_R$  &  & $52.46$ / $0.998$ / $0.471\%$               & $52.43$ / $0.998$  / $\mathbf{0.446}\%$                  & $52.47$ / $0.998$ / $\mathbf{0.449}\%$  &   -           \\
$\FH_D$ &  & $52.55$ / $0.998$  / $\mathbf{0.458\%}$              & $52.53$ / $0.998$ / $0.462\%$  & $52.56$ / $0.998$ / $0.462\%$      & $\mathbf{52.39}$ / $\mathbf{0.998}$ / $\mathbf{8.847}\%$        \\ \bottomrule
\end{tabular}

}
\vspace{-10pt}
\end{table*}

The achieved results of image processing and \gls{sar} data compatibility metrics are depicted in Table~\ref{tab:result_quality_despeckling}.
\rev{\gls{psnr}} and \gls{mssim} do not show significant differences between the possible scenarios: in all situations, \rev{\gls{psnr}}  and \gls{mssim} exceed \rev{$50$~dB} and $0.997$ respectively, meaning for excellent visual quality. 
Nonetheless, omitting the despeckling step allows to achieve the best performance.

A more discriminant metrics 
is the $|\Delta \text{-}\textrm{ENL}|$. 
Excluding the despeckling phase
reveals paramount for keeping at bay the differences in \gls{enl} between attacked and pristine images. 
As anticipated in Section~\ref{subsec:method_despeckling}, despeckling is removing too many high frequency details from the image under attack, including original point scatterers as well\footref{footnote_supp_generic}. On the contrary, simulating a re-acquisition of the amplitude image as it is (i.e., keeping the original speckle noise) allows to reduce the $\textrm{ENL}$ and renders it more similar to the pristine one. 
For this reason, we omit the despeckling step in all the following experiments. 

Another interesting point is that the direct strategy for estimating the frequency response of the system proves to be the best option in terms of \gls{sar} data compatibility. This is worth of notice, considering that this strategy does not need the iterative least squares fitting, thus being less computation demanding. 


\textbf{Effects of the SAR system knowledge level. }
We now evaluate how the quality of attacked images changes according to the attacker knowledge on the original \gls{sar} system frequency response (see Section~\ref{subsubsec:method_sar_simulation}). 
As done before, we omit the best knowledge case (i.e., when $\FH$ is fully known), but we consider four different scenarios in any case: (i) the original \textit{complex} image $\Ic$ under attack is available; (ii) another \textit{complex} image $\mathbf{I}_i e^{j \mathbf{\Phi}_i} \ne \Ic$ captured by the same satellite is available (randomly selected from one different ST-SSC product); (iii) $N$ \textit{complex} images $\{\mathbf{I}_i e^{j \mathbf{\Phi}_i} \}_i^N \ne \Ic $ captured by the same satellite are available; 
in our case, we consider $N = 23$, as we dispose of $24$ ST-SSC products, thus we pick one different image per product; (iv) only the original \textit{amplitude} image $\I$ under attack is available. 

The achieved results are depicted in Table~\ref{tab:result_quality_filter_estimation}\footref{footnote_supp_generic}.
The three different estimation strategies (i.e., $\FH_G$, $\FH_R$ and $\FH_D$) confirm to achieve very similar results among themselves. 
If only the original amplitude image is available, the iterative least square solver did not converge in case of $\FH_G$ and $\FH_R$. This is due to the strongly different frequency content of the amplitude images with respect to the Gaussian and Raised Cosine functions\footref{footnote_supp_generic}.

The $|\Delta \text{-}\textrm{ENL}|$ confirms to be the most discriminant metrics to describe the achieved image quality.
The only situation which reports lower performances is that related with the original amplitude image. This was expected, as the amplitude spectrum presents much lower frequency behaviour with respect to the complex one, thus a consistent portion of high frequency details is filtered out, resulting in an increased $\textrm{ENL}$\footref{footnote_supp_generic}. 
Interestingly, $|\Delta \text{-}\textrm{ENL}|$ does not change consistently if the original complex image is not available, but one or more of other complex images captured by the same satellite can be found. 
The achieved results lead to the conclusion that our counter-forensic attack is valid even in cases of reduced satellite information. One complex image captured by the same satellite is enough to return extremely good performances in reproducing the \gls{sar} acquisition characteristics.



\textbf{Selected attack configuration. }
In light of the previous considerations, we omit the despeckling step and we simulate the re-acquisition by using the direct system estimation strategy with phase-only speckle injection in all the following experiments. 
This option allows to avoid the least squares fitting in the system frequency estimation step and to keep at bay the \gls{enl}-related artifacts on the attacked images.
Moreover, to simplify the scenario, we consider having available the original complex image at attacker side, exploiting it to estimate the \gls{sar} system frequency response. As shown before, this option does not lead to limitations in the presented results.

Some examples of our attack applied to locally manipulated amplitude images are shown in the last column of Fig.~\ref{fig:setup_landcover_forgeries}\footref{footnote_supp_examples}. 
Notice the subtle pattern difference between the manipulated images not passed from our attack (third column) and those undergone the proposed attack. 
This effect is noticeable especially in the splicing area, which results more realistic in terms of speckle noise in the attacked manipulated versions with respect to the manipulated-only versions. 
It is also worth noticing that the semantic content related to the reflectivity scene is maintained. All these characteristics make our attack a valid approach to return high quality realistic images. 

\vspace{-5pt}
\subsection{Counter-forensic Attack Evaluation}
\label{subsec:results_attack}

\begin{table*}[t]
\rev{
\caption{\rev{Average AUC, BA, F1 and IoU in absence of counter-forensic attack. In bold, the \textit{highest} metric achieved per editing operation.} }
\label{tab:result_attack_nospeckle}
\centering
\resizebox{\textwidth}{!}{
\begin{tabular}{lccccccc}
\toprule
&               DownScaleFar &             DownScaleNear &         GaussianBlur-0.5 &                RotateFar &               RotateNear &               UpScaleFar &              UpScaleNear \\
 \midrule
 & AUC / BA / F1 / IoU & AUC / BA / F1 / IoU & AUC / BA / F1 / IoU & AUC / BA / F1 / IoU & AUC / BA / F1 / IoU & AUC / BA / F1 / IoU & AUC / BA / F1 / IoU \\
\midrule
SAE &  $0.98 / 0.97 / 0.92 / 0.89$ &  $1.0 / 0.99 / 0.97 / 0.94$ &  $1.0 / 1.0 / 0.99 / 0.97$ &  $1.0 / 1.0 / 0.98 / 0.97$ &  $1.0 / 1.0 / 0.98 / 0.97$ &  $1.0 / 1.0 / 0.99 / 0.98$ &  $1.0 / 1.0 / 0.99 / 0.98$ \\
ASAE &  $0.99 / 0.98 / 0.94 / 0.91$ &  $0.99 / 0.99 / 0.97 / 0.95$ &  $\mathbf{1.0 / 1.0 / 0.99 / 0.98}$ &  $\mathbf{1.0 / 1.0 / 0.99 / 0.98}$ &  $\mathbf{1.0 / 1.0 / 0.99 / 0.98}$ &  $\mathbf{1.0 / 1.0 / 1.0 / 0.99}$ &  $\mathbf{1.0 / 1.0 / 0.99 / 0.99}$ \\
TruForS &  $\mathbf{1.0 / 1.0 / 0.96 / 0.93}$ &  $\mathbf{1.0 / 1.0 / 0.97 / 0.95}$ &  $1.0 / 1.0 / 0.98 / 0.97$ & $ 1.0 / 1.0 / 0.98 / 0.96$ &  $1.0 / 1.0 / 0.98 / 0.96$ &  $1.0 / 1.0 / 0.99 / 0.98$ &  $1.0 / 1.0 / 0.99 / 0.97$ \\
\midrule
Mean &  $0.99 / 0.98 / 0.94 / 0.91$ &  $1.0 / 0.99 / 0.97 / 0.95$ &  $1.0 / 1.0 / 0.99 / 0.97$ &  $1.0 / 1.0 / 0.98 / 0.97$ &  $1.0 / 1.0 / 0.98 / 0.97$ &  $1.0 / 1.0 / 0.99 / 0.98$ &  $1.0 / 1.0 / 0.99 / 0.98$ \\
\bottomrule
\end{tabular}
}
}
\vspace{-8pt}
\end{table*}

\begin{table*}[t]
\rev{
\caption{\rev{Average AUC, BA, F1 and IoU in case of counter-forensic attack. In bold, the \textit{lowest} metric achieved per editing operation.} }
\label{tab:result_attack_speckle}
\centering
\resizebox{\textwidth}{!}{
\begin{tabular}{lccccccc}
\toprule
&               DownScaleFar &             DownScaleNear &         GaussianBlur-0.5 &                RotateFar &               RotateNear &               UpScaleFar &              UpScaleNear \\
 \midrule
 & AUC / BA / F1 / IoU & AUC / BA / F1 / IoU & AUC / BA / F1 / IoU & AUC / BA / F1 / IoU & AUC / BA / F1 / IoU & AUC / BA / F1 / IoU & AUC / BA / F1 / IoU \\
\midrule
SAE &  $0.8 / 0.78 / 0.46 / 0.40$ &  $0.76 / 0.74 / 0.37 / 0.32$ &  $0.79 / 0.77 / 0.44 / 0.39$ &  $0.79 / 0.77 / 0.44 / 0.38$ &  $0.78 / 0.76 / 0.42 / 0.37$ &  $0.88 / 0.87 / 0.67 / 0.61$ &  $0.81 / 0.79 / 0.49 / 0.43$ \\
ASAE &  $\mathbf{0.69 / 0.68} / 0.29 / 0.24$ &  $\mathbf{0.68 / 0.68} / 0.27 / 0.24$ &  $\mathbf{0.72 / 0.71} / 0.35 / 0.31$ & $ \mathbf{0.71 / 0.7} / 0.33 / 0.29$ &  $\mathbf{0.71 / 0.71} / 0.34 / 0.3$ &  $\mathbf{0.81 / 0.80} / 0.54 / 0.49$ &  $\mathbf{0.75 / 0.74} / 0.41 / 0.37$ \\
TruForS &  $0.77 / 0.72 / \mathbf{0.22 / 0.15}$ & $ 0.74 / 0.70 / \mathbf{0.20 / 0.13}$ &  $0.78 / 0.73 / \mathbf{0.24 / 0.16}$ & $ 0.78 / 0.73 / \mathbf{0.24 / 0.16}$ &  $0.76 / 0.72 / \mathbf{0.23 / 0.15}$ & $ 0.89 / 0.84 / \mathbf{0.47 / 0.35}$ & $ 0.81 / 0.76 / \mathbf{0.29 / 0.20}$ \\
\midrule
Mean &  $0.75 / 0.73 / 0.32 / 0.26$ &  $0.73 / 0.71 / 0.28 / 0.23$ &  $0.76 / 0.74 / 0.34 / 0.29$ &  $0.76 / 0.73 / 0.34 / 0.28$ &  $0.75 / 0.73 / 0.33 / 0.27$ &  $0.86 / 0.84 / 0.56 / 0.48$ &  $0.79 / 0.76 / 0.40 / 0.33$ \\
\bottomrule
\end{tabular}
}
}
\vspace{-8pt}
\end{table*}

\begin{table}[t]
\caption{\rev{Average relative difference of all the metrics (over all editing operations), between the case of counter-forensic attack and the absence of it. In bold, the \textit{worst} case among the three detectors. }}
\label{tab:results_attack_relative_diff}
\centering
\resizebox{.8\columnwidth}{!}{
\rev{
\begin{tabular}{lcccc}
\toprule
& AUC & $\text{BA}$ & $\text{F1}$ & $\text{IoU}$ \\
\midrule
SAE &  $-19.62\%$ & $-21.26\%$ & $-51.78\%$ & $-56.77\%$ \\
ASAE &  $\mathbf{-27.37\%}$ & $\mathbf{-27.99\%}$ & $-63.27\%$ & $-67.09\%$ \\
TruForS &  $-21.0\%$ & $-25.71\%$ & $\mathbf{-72.46\%}$ & $\mathbf{-80.72\%}$ \\
\midrule
Mean &  $-22.66\%$ & $-24.99\%$ & $-62.50\%$ & $-68.19\%$ \\
\bottomrule
\end{tabular}
}
}
\end{table}

We now evaluate the performance of the forensic detectors in presence of our proposed counter-forensic attack.

\textbf{Tampering localization metrics. }
For each analyzed image, we evaluate the tampering localization performances by comparing the detectors output \rev{heatmap} \rev{($\Pbar$ or $\Ptilde$ in absence or presence of attack, respectively)} and the groundtruth tampering mask $\M$.
In particular, we rely on \rev{several metrics: the \gls{auc}, the \gls{ba}, the \gls{f1} and the \gls{iou}. For the metrics requiring a thresholding operation on the heatmap (i.e., the \gls{ba}, the \gls{f1} and the \gls{iou}), we always select their maximum value. We do this to evaluate the detectors in their best possible conditions. 
}
In absence of counter-forensic attacks, we expect \rev{all metrics} to approach $1$, meaning for high detectors accuracy in spotting local manipulations. 
\rev{Our goal is to reduce them, ideally achieving \gls{auc} and \gls{ba} around $0.5$ and \gls{iou} and \gls{f1} around $0$, which means the detectors are not anymore able to distinguish between pristine and manipulated pixels.} 


\textbf{Detectors results in absence of attack. }
To give an upper bound on the achievable detection performances, we first provide the results on the manipulated images not undergone to any counter-forensic attacks. 
Table~\ref{tab:result_attack_nospeckle} depicts the achieved results, averaged over the test dataset, by varying the editing operation applied to the donor image. 
The \rev{metrics associated with all editing operations always achieve high scores, especially for the \gls{asae} detector, which is known to be more accurate than the \gls{sae}~\cite{cannas2022amplitude}, and for the \gls{truforsar} detector}.  

\textbf{Detector results in case of attack. }
Table~\ref{tab:result_attack_speckle} depicts the average detection results in case of testing manipulated amplitude images undergone the proposed counter-forensic attack.
\rev{It is worth noticing an important performance degradation in all scenarios. 
To clarify, we also report in Table~\ref{tab:results_attack_relative_diff} the relative metrics difference with respect to the absence of attack, averaged across all the editing operations. 
On average, the \gls{auc} and the \gls{ba} drop by $\sim 25\%$, while the \gls{f1} and the \gls{iou} by more than $60\%$.
We can observe another interesting fact: the most sensitive detectors to the attack are the \gls{asae} and the \gls{truforsar}, which are actually the best performing detectors in standard situations}. 
This implies that, if the forensic analyst is not aware of potential global attacks done to hide the manipulation area, it is quite probable that the selected detector \rev{will be either the \gls{asae} or the \gls{truforsar}} and that localization results in case of attack will be worse than those achieved by the \gls{sae}. 
Nonetheless, even in case of \gls{sae} detector, we report a strong performance degradation with respect to normal conditions.

\begin{figure}[t]
    \begin{subfigure}[b]{\columnwidth}
        \centering
        \includegraphics[width=\columnwidth]{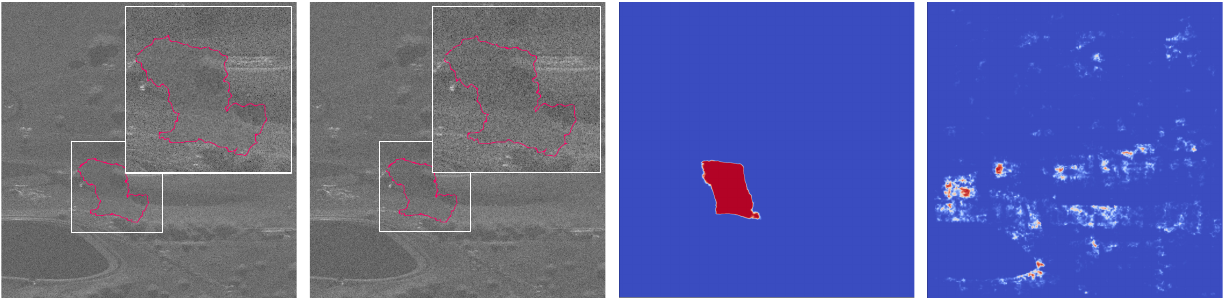}
        \caption{\rev{\scriptsize{\textit{Before}: AUC/BA/F1$=0.97/0.96/0.89$. \textit{After}: AUC/BA/F1$=0.65/0.62/0.09$.}}}     
        \label{fig:results_example_easae_2}
    \end{subfigure}
\hfill
    \begin{subfigure}[b]{\columnwidth}
        \centering
        \includegraphics[width=\columnwidth]{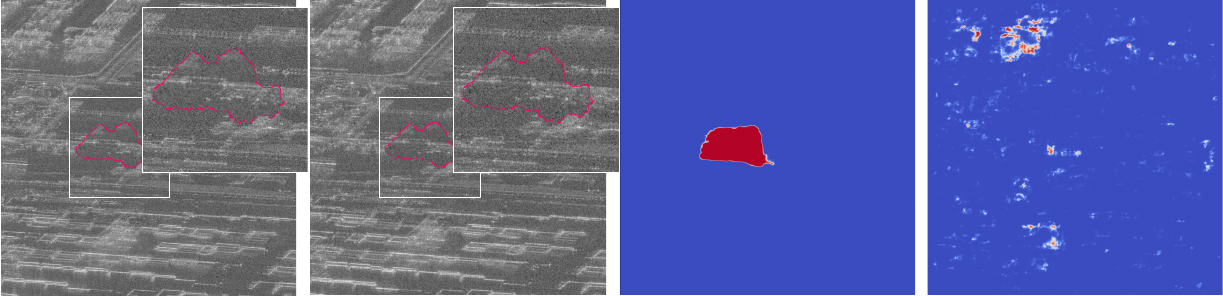}
        \caption{\rev{\scriptsize{\textit{Before}: AUC/BA/F1$=0.94/0.94/0.90$. \textit{After}: AUC/BA/F1$=0.67/0.63/0.08$.}}}
        \label{fig:results_example_easae_4}
    \end{subfigure}
\hfill
    \begin{subfigure}[c]{\columnwidth}
        \centering
        \includegraphics[width=\columnwidth]{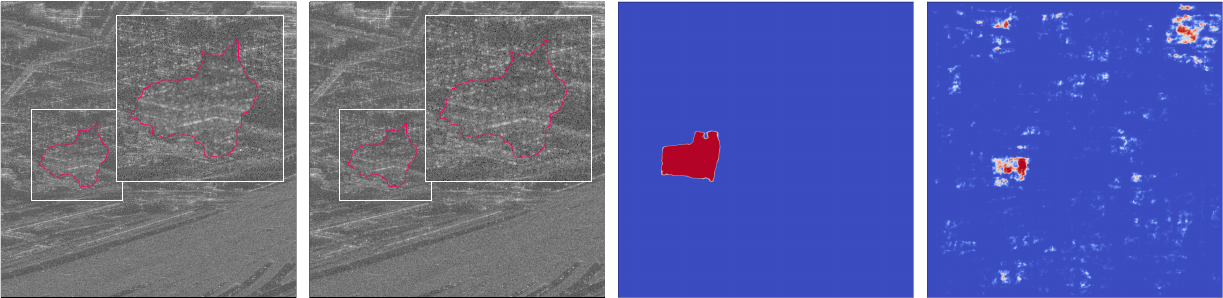}
        \caption{\rev{\scriptsize{\textit{Before}: AUC/BA/F1$=0.94/0.95/0.91$. \textit{After}: AUC/BA/F1$=0.72/0.64/0.24$.}}}
        \label{fig:results_example_easae_3}
    \end{subfigure}
\caption{Examples of the proposed counter-forensic attack applied to manipulated amplitude images. From left to right: the manipulated image in absence of counter-forensic attack; the manipulated image in presence of counter-forensic attack; the \rev{TruForS detector heatmap} in absence of counter-forensic attack; the \rev{TruForS detector heatmap} in presence of counter-forensic attack. 
}   
\label{fig:results_example_asae}
\vspace{-10pt}
\end{figure}

Examples of manipulated attacked images with manually made splicing and related \rev{detection heatmaps} are shown in Fig.~\ref{fig:results_example_asae}\footref{footnote_supp_examples}.
Notice that, even in cases where the \rev{\gls{auc} and \gls{ba} are not perfectly $0.5$ but instead approach $0.7/0.8$, the detector's heatmaps} are considerably less informative than those before the attack. \rev{As we can inspect, the extracted heatmaps in case of attack are not only less informative in the area of tampering, but are also more prone to false alarms, a fact that can affect even more the localization performance.}


\vspace{-20pt}
\rev{\subsection{Comparison with state of the art}}
\label{subsec:sota}
\rev{In this section, we perform further experiments to validate the need of the proposed speckle injection step composing our counter-forensic pipeline. Specifically, we compare our attack with another counter-forensic solution that has been successfully employed to fool tampering localization detectors on natural images~\cite{tailanian2024diffusion}. 
This method is based on the hypothesis that high-frequency patterns carry the most significant forensic traces. Consequently, it focuses on removing high-frequency traces, effectively acting as a denoising step to deceive forensic detectors.}

\rev{
We select this methodology based on the following considerations: (i) it does not require a dedicated training phase, unlike~\cite{cao2024transferable, fang2023attacking}; (ii) it does not rely on prior knowledge of the targeted detector’s functioning, in contrast to~\cite{cao2024transferable, fang2023attacking, boato2024adversarial, rozsa2020adversarial}; (iii) it does not require access to the distribution of pristine samples, unlike~\cite{fang2023attacking, rozsa2020adversarial}; (iv) it does not explicitly necessitate an adaptation phase for \gls{sar} data. These characteristics enable a direct comparison with our proposed attack without requiring modifications.
Furthermore, this method has demonstrated highly promising counter-forensic effects against the TruFor detector on natural images, reducing its localization performance by more than $24\%$ across various datasets~\cite{tailanian2024diffusion}.}

\rev{To evaluate its effectiveness, we first verify the quality of attacked images, then, we evaluate its counter-forensic effect.}

\rev{\textbf{Quality of the attacked images}. 
We select the same set of pristine images used for the evaluations done in Section~\ref{subsec:results_quality} and we attack them with the method proposed in~\cite{tailanian2024diffusion}. The achieved results show an average $\textrm{PSNR}=53.15 \, \textrm{dB}, \textrm{MS-SSIM}=0.999$ and $|\Delta \text{-}\textrm{ENL}| = 2.075\%$. These numbers should be compared with those achieved by our selected attack configuration, shown in Table~\ref{tab:result_quality_filter_estimation}.}

\rev{
Notably, the method proposed in~\cite{tailanian2024diffusion} provides better \gls{psnr} and \gls{mssim} than our proposed attack, even though the results are comparable.
Nonetheless, as it was already pointed out, a more discriminant metrics for \gls{sar}-related data is the $|\Delta \text{-}\textrm{ENL}|$, being the \gls{enl} a statistical measure to quantify the level of speckle noise in images. 
Our proposed approach returns a $|\Delta \text{-}\textrm{ENL}| = 0.458\%$, enabling to better reduce the difference between the \gls{enl} of attacked images and that of pristine ones. We believe that method~\cite{tailanian2024diffusion}, being focused on removing high-frequency traces and actually acting as a denoising step, partially fails in reproducing the unique granular properties of the speckle noise. }

\rev{\textbf{Counter-forensic attack evaluation}. 
Table~\ref{tab:comparison_sota_counter_forensics_sae} compares the average localization results after the counter-forensic attack developed in~\cite{tailanian2024diffusion} and our proposed attack. 
As we can quickly observe, the attack proposed in~\cite{tailanian2024diffusion} does not effectively hide manipulation traces in \gls{sar} images. Indeed, the difference between localization metrics before and after the attack is negligible. In contrast, our proposed methodology effectively conceals these traces, leading to a significant performance drop across all metrics.}

\begin{table}[t]
\caption{\rev{Localization results, averaged over all detectors (SAE, ASAE and TruForS) and over all editing operations, achieved on images attacked with the method proposed in~\cite{tailanian2024diffusion} and with our proposed method, in terms of average AUC, BA, F1 and IoU. We show results after the attack and the relative difference with respect to the absence of attack ($\Delta \%$). In bold, the best counter-forensic results.}}
\label{tab:comparison_sota_counter_forensics_sae}
\centering
\resizebox{\columnwidth}{!}{
\rev{
\begin{tabular}{lcccc}
\toprule
& AUC & $\text{BA}$ & $\text{F1}$ & $\text{IoU}$ \\
\midrule
 & \text{attack / } $\Delta\% $& \text{attack / } $\Delta\%$ & \text{attack / } $\Delta\%$ & \text{attack / } $\Delta\%$ \\
\midrule
\text{\cite{tailanian2024diffusion}} & \(0.99/-0.53\%\) & \(0.99/-0.63\%\) & \(0.96/-1.86\%\) & \(0.94/-2.51\%\) \\
\text{Ours} & $\mathbf{0.77/-22.66\%}$ & $\mathbf{0.75/-24.99\%}$ & $\mathbf{0.37/-62.50\%} $ & $\mathbf{0.31/-68.19\% }$ \\
\bottomrule
\end{tabular}
}
}
\vspace{-10pt}
\end{table}

\vspace{5pt}
\subsection{Realistic Attacks}

In this section, we test the proposed methodology on particular cases which resemble even more challenging real-world attacks. 
Indeed, the previous analysis allows a large number of
precisely controlled experiments, though squared forgery areas 
which have been automatically generated
might be little realistic.

We consider an additional dataset including \gls{sar} data known as \gls{mstar}, which contains public X-band complex \gls{sar} images of 
military tank targets at $1$ foot resolution ($\sim 30\textrm{cm}$) \cite{mstar_dataset}. 
We create local manipulations by inserting target vehicles in \gls{sar} amplitude images selected from our TerraSAR-X testing dataset.
Given that the average size of a military tank is at maximum $10 \times 4$ meters in length and width, we consider donor pixel areas around $35 \times 16$ pixels. This size is not fixed as it depends on the specific target vehicle to insert. 

To create the attack, we follow the pipeline described in Section~\ref{sec:method}, including phase-only speckle noise injection and estimating the frequency response of the system directly from the available original complex image. 

\begin{figure}[t]
\centering
    \begin{subfigure}[b]{\columnwidth}
        \centering
        \includegraphics[width=\columnwidth]{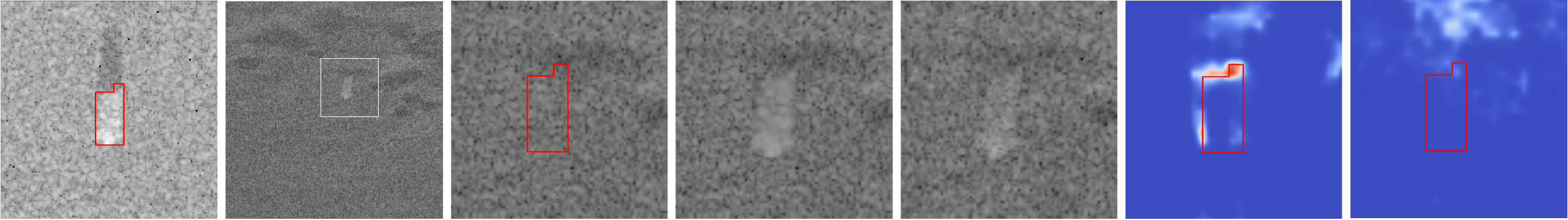}
        \label{fig:results_mstar1}
        \vspace{-11pt}
    \end{subfigure}
\hfill
    \begin{subfigure}[b]{\columnwidth}
        \centering
        \includegraphics[width=\columnwidth]{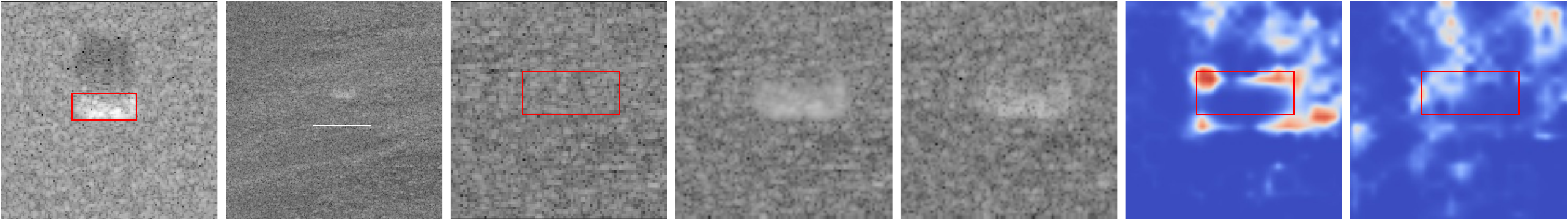}
        \label{fig:}
                \vspace{-11pt}
    \end{subfigure}
\hfill
    \begin{subfigure}[c]{\columnwidth}
        \centering
        \includegraphics[width=\columnwidth]{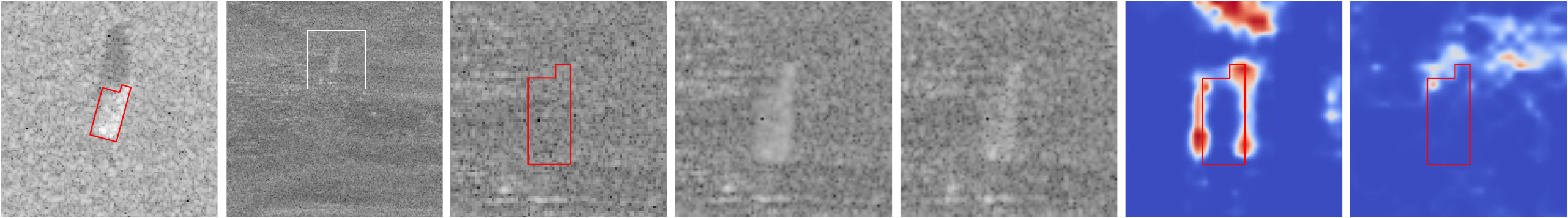}
        \label{fig:}
    \end{subfigure}
\caption{Examples of the proposed counter-forensic attack applied to manipulated amplitude images by exploiting military vehicles selected from MSTAR dataset. From left to right: the donor SAR amplitude image of the vehicle; the manipulated image in presence of attack; close-up of the pristine image of the area inside the white box; close-up of the manipulated image in absence of attack; close-up of the manipulated image in presence of attack; close-up of the \rev{TruForS detector heatmap} in absence of attack; close-up of the \rev{TruForS detector heatmap} in presence of attack. 
}   
\label{fig:result_realistic_example}
\vspace{-10pt}
\end{figure}

Some splicing examples and their related detector's \rev{heatmaps} are shown in Fig.~\ref{fig:result_realistic_example}\footref{footnote_supp_examples}.
\rev{In absence of attack, the detectors are still able to localize the area of tampering (at least in its contours), even if they struggle a little bit (see the sixth column). We believe this is due to the extremely small manipulation area, which proves quite challenging to be exposed even for the most advanced localization methods.}

Notice that the attacked spliced images still carry semantic traces of the inserted vehicle: in correspondence of the tank position, the scattered amplitude has a visible peak (see the fifth column). Despite this fact, the detectors are blind to the attacked image, and \rev{their heatmap} cannot distinguish the splicing area from the surrounding pixels. 

\vspace{-5pt}
\subsection{Results' Discussion}

We believe the strength of the proposed approach lies in the exploitation of the complex nature of \gls{sar} images for hindering the manipulation traces. 
Indeed, the speckle injection step contributes in widening the frequency spectrum of the manipulated image and conceals the subtle editing traces searched by forensic detectors.
\rev{As a matter of fact, counter-forensic solutions available in state of the art~\cite{tailanian2024diffusion} or other global editing operations\footref{footnote_supp_generic} do not apply processing on complex images, as they act only on the magnitude component.} 

The further filtering operation through the estimated end-to-end \gls{sar} system provides the correct frequency shaping to reproduce the same \gls{sar} characteristics of original pristine images. 
This step is also quite important for hiding local editing artifacts that would be noticeable at human inspection even after the speckle injection stage. 
To reproduce exactly the same pattern of the original \gls{sar} images, we have to apply the correct shaping filter corresponding to the \gls{sar} system used for pristine acquisitions\footref{footnote_supp_generic}.



\section{Conclusions}
\label{sec:conclusions}

In this paper, we prove that an experienced \gls{sar} practitioner can effectively conceal local manipulation traces within \gls{sar} amplitude images. 
By exploiting the complex nature of \gls{sar} data, our devised counter-forensic attack involves simulating a re-acquisition of the manipulated amplitude image by the same \gls{sar} system that was used to acquire its original, unaltered, complex version. 
Through our experimental campaign, we show how the attacker can obscure all signs of manipulation, thereby presenting the image as if it were legitimately acquired by the system.

We believe our proposed methodology holds significant advantages for the \gls{sar} forensic community for three primary reasons: first, it allows to simulate worrying scenarios in which a malicious user aims at erasing the traces of local manipulations; second, it answers the needs for privacy preservation by hiding sensitive targets; finally, it assesses the robustness of existing forensic detectors in case of adversarial attacks. 

We evaluate the efficacy of the suggested counter-forensic method across a range of scenarios, analyzing different manipulation techniques. We experiment on data undergone to automatic and hand-made local manipulations, considering also the alarming scenario of military vehicles insertion. 
The results obtained demonstrate that our developed attack effectively eradicates any evidence of manipulation, even fooling sophisticated forensic detection systems. 
Moreover, the visual quality and the \gls{sar} properties of the attacked images are maintained. 
\rev{We also compare the proposed attack with another counter-forensic solution available in the literature. Our attack demonstrates better reconstruction properties in terms of \gls{sar}-related features and significantly stronger anti-forensic capabilities. }


\rev{\textbf{Possible Future Directions}. 
In our proposed work, we limit the attack to concealing manipulation traces by injecting speckle noise from the same \gls{sar} system used for acquiring the original images. However, realistic scenarios could include situations where attackers might need to disguise a manipulated image as a pristine one captured by \textit{another} \gls{sar} system. We believe that, in principle, our proposed methodology would still work, given that the other satellite chosen as the ``target'' system has very similar characteristics (acquisition mode, processing pipeline, etc.) with respect to the original one used for acquisitions. We reserve the possibility of exploring this in future studies.
}

\rev{Another interesting aspect to investigate is the transferability of the proposed attack to other detection models, such as \gls{sar} \gls{atr} models, which are commonly used to extract remote sensing information and have already revealed several vulnerabilities to adversarial attacks~\cite{du2021adversarial,ye2023realistic}. Future works will focus on exploring whether our counter-forensic attack, designed for localization detectors, can also be effective in deceiving \gls{atr} models.}

\rev{The security implications of our proposed attack highlight the urgent need to discuss potential countermeasures and future improvements for forensic detectors.
For example, if forensic analysts know in advance the nature of the counter-forensic attack, they could include attacked manipulated data during the training phase of forensic detectors.
By doing so, the detectors could potentially learn to extract a meaningful localization heatmap even if the manipulated data had undergone our proposed attack.
Although this countermeasure can enhance robustness to the attack\footref{footnote_supp_generic}, it does not fully resolve the challenges associated with the removal of manipulation traces, an issue that should be thoroughly addressed in future research.}




\ifCLASSOPTIONcaptionsoff
  \newpage
\fi

\bibliographystyle{IEEEtran}
\bibliography{biblio}



\end{document}